\documentclass{article}

\usepackage{PRIMEarxiv}

\usepackage[utf8]{inputenc} % allow utf-8 input
\usepackage[T1]{fontenc}    % use 8-bit T1 fonts
\usepackage{hyperref}       % hyperlinks
\usepackage{url}            % simple URL typesetting
\usepackage{booktabs}       % professional-quality tables
\usepackage{amsfonts}       % blackboard math symbols
\usepackage{nicefrac}       % compact symbols for 1/2, etc.
\usepackage{microtype}      % microtypography
\usepackage{lipsum}
\usepackage{fancyhdr}       % header
\usepackage{graphicx}       % graphics
\graphicspath{{media/}}     % organize your images and other figures under media/ folder

\usepackage{natbib}
\usepackage{amsmath} 
\usepackage{amssymb}
\usepackage{xcolor}
\usepackage{subcaption}
\usepackage{tabularx}
\usepackage{multirow}
\usepackage{algorithm}
\usepackage{algpseudocode}
\usepackage{comment}
\usepackage{float}
\usepackage{booktabs}
\usepackage{bm}  
\usepackage{placeins}

\title{Soft Partition-based KAPI-ELM for Multi-Scale PDEs}

\author{
  Vikas Dwivedi\thanks{Corresponding Author} \\
  CREATIS Biomedical Imaging Laboratory \\
  INSA, CNRS UMR 5220, Inserm, Universit´e Lyon 1 \\
  Lyon 69621, France\\
  \texttt{vikas.dwivedi@creatis.insa-lyon.fr} \\
     \And
  Monica Sigovan \\
  CREATIS Biomedical Imaging Laboratory \\
  INSA, CNRS UMR 5220, Inserm, Universit´e Lyon 1 \\
  Lyon 69621, France\\
  \texttt{monica.sigovan@insa-lyon.fr} \\
    \And
  Bruno Sixou \\
  CREATIS Biomedical Imaging Laboratory \\
  INSA, CNRS UMR 5220, Inserm, Universit´e Lyon 1 \\
  Lyon 69621, France\\
  \texttt{bruno.sixou@insa-lyon.fr} \\
}

\begin{document}
\maketitle

\begin{abstract}
	Physics-informed machine learning holds great promise for solving differential equations, yet existing methods struggle with highly oscillatory, multiscale, or singularly perturbed PDEs due to spectral bias, costly backpropagation, and manually tuned kernel or Fourier frequencies. This work introduces a soft partition--based Kernel-Adaptive Physics-Informed Extreme Learning Machine  (KAPI--ELM), a deterministic low-dimensional parameterization in which smooth partition lengths jointly control collocation centers and Gaussian kernel widths, enabling continuous coarse-to-fine resolution without Fourier features, random sampling, or hard domain interfaces. A signed-distance–based weighting further stabilizes least-squares learning on irregular geometries. Across eight benchmarks—including oscillatory ODEs, high-frequency Poisson equations, irregular-shaped domains, and stiff singularly perturbed convection–diffusion problems—the proposed method matches or exceeds the accuracy of state-of-the-art Physics-Informed Neural Network (PINN) and Theory of Functional Connections (TFC) variants while using only a single linear solve. Although demonstrated on steady linear PDEs, the results show that soft-partition kernel adaptation provides a fast, architecture-free approach for multiscale PDEs with broad potential for future physics-informed modeling. For reproducibility, the reference codes are available at~\url{https://github.com/vikas-dwivedi-2022/soft_kapi}
\end{abstract}

% keywords can be removed

\section{Introduction}
\label{Sec:Introduction}

Physics-informed machine learning has become a widely used paradigm for solving differential equations across science and engineering, offering mesh-free formulations that embed physical laws directly into the learning process.  
However, many real-world models exhibit \emph{multiscale structure}, \emph{highly oscillatory dynamics}, or \emph{stiff boundary layers}—regimes in which existing approaches often become inaccurate, unstable, or computationally expensive.

Physics-Informed Neural Networks (PINNs~\citep{RAISSI2019686}) have achieved impressive results on smooth PDEs, but they suffer from spectral bias~\citep{WANG2022110768,pmlr-v97-rahaman19a,CiCP-28-5}, slow gradient-based training~\citep{10.1093/imamat/hxae011}, and strong sensitivity to architectural choices for PDEs with sharp gradients~\citep{ABBASI2025131440, Luo2025}.  
Fourier-feature PINNs~\citep{NEURIPS2020_55053683} and sinusoidal embeddings partially mitigate high-frequency underfitting, yet introduce additional frequency hyperparameters and lead to high computational cost~\citep{Moseley2023}.  
Domain-decomposition PINNs~\citep{Klawonn2024} improve local resolution but require \emph{hard interface constraints}, complicating the loss landscape and becoming fragile for extremely thin boundary layers~\citep{KO2025113860,ARZANI2023111768,Gie21092024}.  
Physics-Informed Extreme Learning Machines (PI--ELMs~\citep{DWIVEDI202096,DWIVEDI_DPINN_2021,DONG2021114129}) remove backpropagation entirely and train through a single least-squares solve, but their \emph{randomly sampled} Gaussian centers and widths cannot adapt to localized multiscale features~\citep{DWIVEDI2025130924}.  
Recent extensions using Fourier feature mappings~\citep{ren2025generalfourierfeaturephysicsinformed} improve high-frequency accuracy but again depend on manual frequency tuning.  
Kernel-Adaptive PI--ELM (KAPI--ELM~\citep{dwivedi2025kerneladaptivepielmsforwardinverse}) introduced distributional hyperparameters to adapt sampling, yet the method remains slightly overparameterized, heuristic-driven in terms of RBF width decay rate, and has not been systematically extended to a general multiscale-cum-singularly perturbed PDE solver.

Despite significant progress, there remains no simple adaptive sampling strategy that can:
(i) simultaneously resolve high-frequency oscillations and exponentially thin boundary layers within a single formulation,  
(ii) avoid Fourier feature mappings and their frequency-tuning burden,  
(iii) eliminate interface penalties required by domain-decomposition approaches, and  
(iv) retain the analytic efficiency and speed of PI--ELM.  
Existing methods rely either on heavy backpropagation (PINNs, Fourier-PINNs), manual frequency engineering (Fourier-feature PINNs/ELMs), or stochastic and heuristic distributional tuning (original KAPI--ELM).  
None of them provide a unified geometric mechanism that \emph{smoothly} allocates resolution across the domain, especially for multiscale PDEs where layer thickness can vary by orders of magnitude.

To bridge these gaps, we introduce a \emph{soft partition–based Kernel-Adaptive PI--ELM} (KAPI--ELM) framework that resolves multiscale, oscillatory, and singularly perturbed PDEs without Fourier embeddings, neural-network architectures, or domain-decomposition interface losses.  
The core idea is to replace pointwise or stochastic adaptivity with a deterministic \emph{soft partition} of the domain: smoothly varying partition lengths jointly control (i) the density of collocation centers and (ii) the local scaling of Gaussian kernels.  
This yields continuous rather than abrupt refinement, avoids hard interfaces, and provides a natural coarse-to-fine mechanism aligned with boundary layers, interior gradients, and high-frequency modes.  
In parallel, an SDF-weighted PDE residual formulation stabilizes learning near irregular boundaries, enabling high accuracy on complex geometries without meshing or reparameterization.  
The resulting method remains fully backpropagation-free: all solutions are obtained from a single linear least-squares system, and only a handful of hyperparameters (partition lengths) are optimized externally.

\paragraph{Contributions}
The main contributions of this work are as follows:

\begin{itemize}
	
	\item \textbf{Soft partition–based sampling for adaptive multiscale resolution.}  
	We introduce a deterministic, low-dimensional parameterization in which smooth partition lengths govern both the placement of collocation centers and their associated Gaussian widths, producing stable coarse-to-fine adaptivity without hard interfaces.
	
	\item \textbf{Unified framework for oscillatory, multiscale, and singularly perturbed PDEs.}  
	The proposed method is, to our knowledge, the first PI--ELM-based approach capable of resolving high-frequency oscillations and exponentially thin boundary layers within the same formulation, without Fourier feature mappings or domain decomposition.
	
	\item \textbf{Fourier-free high-frequency accuracy.}  
	The method achieves multiscale expressivity purely through analytically prescribed Gaussian widths tied to partition lengths—requiring no Fourier embeddings, frequency heuristics, or manual tuning.
	
	\item \textbf{Geometry-agnostic SDF-weighted residuals.}  
	We introduce a signed-distance–based weighting strategy that stabilizes least-squares learning near irregular boundaries while remaining fully mesh-free and easily applicable to arbitrary domains.
	
	\item \textbf{Efficient, backprop-free, and low-dimensional optimization.}  
	All solutions reduce to a single linear least-squares solve, and the external optimization acts on only a small number of partition parameters, yielding significant speedups over PINNs, Fourier-PINNs, and earlier KAPI--ELM variants.
	
\end{itemize}

\paragraph{Organization}
The remainder of this paper is organized as follows.  
Section~\ref{Sec:Prelim} revisits PI--ELM and the original KAPI--ELM framework.  
Section~\ref{Sec:Sampling} introduces the proposed soft partition–based KAPI--ELM, including center and width construction in one and two dimensions, the associated optimization strategy, and the SDF-weighted residual formulation.  
Section~\ref{Sec:Results} presents extensive numerical experiments—covering highly oscillatory ODEs, high-frequency 2D Poisson problems, irregular-domain elliptic PDEs, and stiff singularly perturbed boundary-layer equations—demonstrating the method's accuracy, robustness, and multiscale capability.  
Section~\ref{sec:limitations} outlines current limitations and opportunities for extending soft partitioning to transient, nonlinear, and higher-dimensional PDEs.  
Finally, Section~\ref{Sec:Conclusion} summarizes the main findings and discusses broader implications for mesh-free physics-informed kernel methods.

\section{Preliminaries: PI--ELM and Kernel-Adaptive PI--ELM}
\label{Sec:Prelim}

We consider a linear differential operator 
$\mathcal{L}_{\nu}(\cdot)$ defined on a domain 
$\Omega \subset \mathbb{R}$ with boundary $\partial\Omega$, 
parameterized by a PDE coefficient $\nu$.
The goal is to approximate the solution of
\begin{equation}
	\mathcal{L}_{\nu}(u) = R(x), \qquad x\in\Omega,
	\qquad
	u(x)=g(x), \qquad x\in\partial\Omega,
	\label{eq:pde}
\end{equation}
where $R$ is the source term and $g$ the boundary data.

\subsection{Physics-Informed Extreme Learning Machine (PI--ELM)}

PI--ELM \citep{DWIVEDI202096,CALABRO2021114188} represents the solution as a global
expansion of $N^*$ Gaussian radial basis functions,
\begin{equation}
	\hat{u}(x)
	= \sum_{i=1}^{N^*} 
	c_i\exp\!\left[-\frac{(x-\alpha_i^*)^2}{2\sigma_i^2}\right],
	\label{eq:pielm_hypothesis}
\end{equation}
where the RBF centers $\alpha_i^*$ and widths $\sigma_i$ are drawn
randomly from prescribed distributions, and the output weights
$\mathbf{c}=[c_1,\ldots,c_{N^*}]^{\top}$ are determined analytically.
Enforcing the PDE residual and boundary conditions at the collocation
points yields a linear system
\begin{equation}
	H\mathbf{c} = \mathbf{r},
	\label{eq:pielm_linear}
\end{equation}
whose minimum-norm least-squares solution is obtained via the
Moore--Penrose pseudoinverse,
\begin{equation}
	\mathbf{c}=H^{\dagger}\mathbf{r}.
	\label{eq:pielm_solution}
\end{equation}
Thus PI--ELM requires no backpropagation and trains in a single
least-squares step.
However, its accuracy depends strongly on the randomly sampled kernel
parameters $(\alpha_i^*,\sigma_i)$, which remain fixed during training
and cannot adapt to sharp layers or multiscale behavior.

\subsection{Kernel-Adaptive PI--ELM (Original KAPI--ELM)}

The original KAPI--ELM \citep{dwivedi2025kerneladaptivepielmsforwardinverse} augments PI--ELM by
introducing a low-dimensional set of \emph{distributional hyperparameters}
that govern the sampling of RBF centers and widths.
Instead of fixing $(\alpha_i^*,\sigma_i)$ in advance, KAPI--ELM defines
a parametric mixture model for the centers and an inverse-scale
distribution for the widths.
For any choice of the hyperparameter vector $\boldsymbol{w}$, the
inner PI--ELM problem \eqref{eq:pielm_linear}--\eqref{eq:pielm_solution}
is solved in closed form, and an outer Bayesian optimization adjusts
$\boldsymbol{w}$ so that the induced kernel distribution adapts to the
solution features.
This improves accuracy over PI--ELM, particularly for problems with
localized gradients or boundary layers, while preserving the analytic
efficiency of the least-squares solve.

While effective, the original KAPI--ELM relies on several
distributional hyperparameters—such as mixture weights, means,
variances, and width-decay factors—whose selection depends on
problem-specific heuristics tied to the PDE parameter.  
This makes the formulation somewhat overparameterized, empirically
driven, and stochastic, with the outer optimization exhibiting
sensitivity to initialization, particularly for multiscale PDEs.  
These observations motivate the more structured and deterministic
parameterization introduced in the next section.

\section{Methodology: Soft Partition-Based KAPI--ELM Framework}

The proposed soft partition-based KAPI–ELM introduces a low-dimensional
\emph{distributional parameterization} that controls both
(1) the placement of collocation centers and
(2) the widths of Gaussian kernel basis functions,
without imposing any hard interfaces between subregions of the domain.
In contrast to classical adaptive schemes—which adjust individual point
locations or basis widths and often create discontinuous refinement zones—the
soft-partition strategy uses a small set of trainable partition lengths to
smoothly modulate center density and kernel scales across the domain.
All collocation locations and widths are deterministic, continuous functions of
these partition parameters, enabling stable optimization, eliminating the
high-dimensional instability of pointwise adaptivity, and providing a natural
coarse-to-fine mechanism for resolving multiscale features.

%-------------------------------------------------------------
\subsection{One-Dimensional Sampling Strategy}
\label{Sec:Sampling}

Consider a 1D computational domain $[0,1]$.  
We introduce a vector of positive partition lengths
\[
\boldsymbol{\ell} = (\ell_1,\ldots,\ell_k),
\qquad
\ell_j > 0,
\qquad
\sum_{j=1}^k \ell_j = 1,
\]
which defines the partition boundaries
\[
0 = x_0 < x_1 = \ell_1 < x_2 = \ell_1 + \ell_2
< \cdots < x_k = 1.
\]
The components of $\boldsymbol{\ell}$ constitute the \emph{only}
trainable hyperparameters of the 1D sampling strategy.
Because the vector is normalized, the number of effective degrees of
freedom equals $k-1$.

%-------------------------------------------------------------
\paragraph{Partition Centers}
Each subinterval $[x_{j-1},x_j]$ receives exactly $N$ collocation
centers,
\[
x^{(\mathrm{part})}_{j,m}
= x_{j-1} + \frac{m}{N}\,\ell_j,
\qquad m=0,\ldots,N,
\]
with duplicates at interior boundaries removed.
Concatenating these yields $kN$ \emph{partition-based centers},
denoted $x_{\mathrm{part}}$.

%-------------------------------------------------------------
\paragraph{Global Centers}
To ensure full-domain coverage regardless of partition lengths, 
we also construct a uniform grid of $kN$ global centers,
\[
x^{(\mathrm{glob})}_r
= \frac{r}{kN+1},
\qquad r = 1,\ldots,kN,
\]
forming the set $x_{\mathrm{glob}}$.

%-------------------------------------------------------------
\paragraph{Width Assignment}

Each center $c_i$ is associated with a Gaussian kernel
\[
\phi_i(x)
= \exp\!\left( -\frac{(x - c_i)^2}{2\sigma_i^2} \right),
\]
where the width $\sigma_i$ is determined analytically by the partition
lengths.

\emph{Partition widths.}
All $N$ centers in subinterval $j$ share the piecewise-constant width
\[
\sigma_j = k_\sigma \frac{\ell_j}{N},
\]
with $k_\sigma > 0$ a global scale parameter.

\emph{Global widths.}
The global centers use a constant width
\[
\sigma_{\mathrm{glob}}
= k_\sigma \frac{1}{kN}.
\]

Thus small partition lengths produce dense centers with narrow kernels,
ideal for resolving steep gradients or boundary layers, while larger
partitions produce coarser sampling appropriate for smooth regions.

%-------------------------------------------------------------
\paragraph{Distributional Manifold}
All centers and widths depend deterministically on
$\boldsymbol{\ell}$, defining a low-dimensional manifold
\[
\mathcal{M}
= 
\bigl\{
(x_{\mathrm{part}}, \sigma_{\mathrm{part}},
x_{\mathrm{glob}}, \sigma_{\mathrm{glob}})
\;\big|\;
\boldsymbol{\ell} \in \Delta_{k-1}
\bigr\}.
\]
Compared with traditional meshfree adaptivity that requires optimizing
hundreds or thousands of point locations, KAPI--ELM optimizes only
$k-1$ variables.

%-------------------------------------------------------------
\paragraph{Coarse-to-Fine Adaptation}
The method typically begins with $k=2$, corresponding to a single
effective trainable parameter.
Increasing $k$ increases expressive power by permitting multiple
localized partitions.
Because the global grid is always present, the method preserves
robustness as resolution increases.

%-------------------------------------------------------------
\subsection{Two-Dimensional Sampling Strategy}

The 1D construction extends naturally to $\Omega = [0,1]^2$.
Let
\[
\boldsymbol{\ell}^x = (\ell^x_1,\ldots,\ell^x_{k_x}),
\qquad
\boldsymbol{\ell}^y = (\ell^y_1,\ldots,\ell^y_{k_y}),
\]
with each vector positive and normalized to unity.
The effective dimension of the parameterization equals
$(k_x-1)+(k_y-1)$.

%-------------------------------------------------------------
\paragraph{Partition-Based Centers}
Partition boundaries along each axis are
\[
x_i = \sum_{r=1}^i \ell^x_r,
\qquad
y_j = \sum_{s=1}^j \ell^y_s,
\]
forming rectangles
\[
\Omega_{ij}
=
[x_{i-1},x_i] \times [y_{j-1},y_j].
\]
Each block $\Omega_{ij}$ receives an $N_x\times N_y$ tensor grid of
points, with interior-duplicate removal ensuring uniqueness, yielding
$k_xk_yN_xN_y$ partition-based centers.

%-------------------------------------------------------------
\paragraph{Partition-Based Widths}
Grid spacings in block $\Omega_{ij}$ are
\[
\Delta x_{ij} = \frac{\ell^x_i}{N_x},
\qquad
\Delta y_{ij} = \frac{\ell^y_j}{N_y},
\]
and the corresponding isotropic Gaussian width is
\[
\sigma_{ij}
= k_\sigma \sqrt{\Delta x_{ij}\,\Delta y_{ij}}.
\]

%-------------------------------------------------------------
\paragraph{Global Centers and Width}
A uniform tensor-product grid with the same cardinality is constructed,
and assigned the constant width
\[
\sigma_{\mathrm{glob}}
= k_\sigma \sqrt{
	\frac{1}{k_x N_x}
	\cdot
	\frac{1}{k_y N_y}
}.
\]

%-------------------------------------------------------------
\paragraph{Distributional Manifold (2D)}
All centers and kernel widths are deterministic functions of
$(\boldsymbol{\ell}^x,\boldsymbol{\ell}^y)$,
yielding a low-dimensional manifold
\[
\mathcal{M}_{2\mathrm{D}}
=
\{
(\mathrm{XY}_{\mathrm{part}}, \sigma_{\mathrm{part}},
\mathrm{XY}_{\mathrm{glob}}, \sigma_{\mathrm{glob}})
\mid 
\boldsymbol{\ell}^x\in\Delta_{k_x-1},
\boldsymbol{\ell}^y\in\Delta_{k_y-1}
\}.
\]

%-------------------------------------------------------------
\subsection{Optimization Strategy}

For any set of partition parameters, the sampling strategy determines
all collocation centers and Gaussian widths.
The PI--ELM coefficients satisfy the least-squares system
\begin{equation}
	H(\boldsymbol{\ell})\,\mathbf{c}
	=
	\mathbf{r},
	\label{eq:inner_ls}
\end{equation}
with closed-form solution
\[
\mathbf{c}^*(\boldsymbol{\ell})
=
H(\boldsymbol{\ell})^{\dagger}\mathbf{r}.
\]

An independent validation grid defines the outer objective
\begin{equation}
	J(\boldsymbol{\ell})
	=
	\bigl\|
	H_{\mathrm{val}}(\boldsymbol{\ell})
	\,\mathbf{c}^*(\boldsymbol{\ell})
	-\mathbf{r}_{\mathrm{val}}
	\bigr\|_{\infty}.
\end{equation}

The optimal partition parameters solve
\[
\boldsymbol{\ell}^{*}
=
\arg\min_{\boldsymbol{\ell}\in\Delta_{k-1}}
J(\boldsymbol{\ell})
\qquad\text{(1D)},
\]
or
\[
(\boldsymbol{\ell}^{x*},\boldsymbol{\ell}^{y*})
=
\arg\min_{(\boldsymbol{\ell}^x,\boldsymbol{\ell}^y)
	\in \Delta_{k_x-1}\times\Delta_{k_y-1}}
J(\boldsymbol{\ell}^x,\boldsymbol{\ell}^y)
\qquad\text{(2D)}.
\]

Because the search space for the partition parameters is extremely low-dimensional, we employ Bayesian optimization with a Gaussian–process surrogate to efficiently minimize the validation objective. The GP surrogate provides an uncertainty-aware approximation of $J(\boldsymbol{\ell})$, and new partition candidates are selected using the expected-improvement (EI) acquisition rule, which balances exploration and exploitation. The simplex constraints on the partition lengths are enforced through a softmax reparameterization, ensuring valid partitions throughout the search. In practice, only 20--40 Bayesian-optimization iterations are sufficient for convergence, making the outer optimization inexpensive relative to the inner least-squares solve.

%-------------------------------------------------------------
\subsection{Boundary-Aware PDE Weighting via Signed Distance Functions}
\label{sec:sdf_weighting}

In irregular domains, uniform weighting of PDE residuals often causes an
imbalance between boundary and interior constraints: PDE rows near 
$\partial\Omega$ may dominate or conflict with Dirichlet/Neumann rows,
leading to boundary leakage, poor conditioning, or unstable least-squares
solves.  
To mitigate this, we introduce a geometry-aware residual weighting strategy 
based on the signed distance function (SDF).

Let
\[
d(\mathbf{x})
=
\mathrm{dist}(\mathbf{x},\partial\Omega),
\qquad 
d(\mathbf{x})>0 \ \text{inside}, 
\quad
d(\mathbf{x})=0 \ \text{on } \partial\Omega .
\]
Each PDE row in the least-squares system is multiplied by a smoothly varying
weight
\begin{equation}
	\label{eq:sdf_weight}
	w_{\mathrm{PDE}}(\mathbf{x})
	=
	w_{\mathrm{near}}
	+
	\bigl(w_{\mathrm{far}} - w_{\mathrm{near}}\bigr)\,
	\rho\!\left( \frac{d(\mathbf{x})}{\delta} \right),
\end{equation}
where $\rho:[0,\infty)\to[0,1]$ is a smooth ramp function (e.g.~cubic or
quintic polynomial) satisfying
\[
\rho(0)=0,
\qquad
\rho(s)=1 \ \text{ for } s\ge 1.
\]
To the best of our knowledge, SDF-based \emph{residual weighting}—as opposed
to SDF-based trial functions~\citep{SUKUMAR2022114333} for exact boundary satisfaction—has not been
explored in the PINN or PI--ELM literature, where distance functions are
typically used only to encode geometry or enforce constraints analytically.

The parameter $\delta>0$ controls the thickness of the boundary transition
region. For $0 \le d(\mathbf{x}) \le \delta$, the weight increases smoothly
from $w_{\mathrm{near}}$ (relaxed PDE enforcement near boundaries) to
$w_{\mathrm{far}}$ (full PDE enforcement in the interior), and for
$d(\mathbf{x})\ge \delta$ one simply has $w_{\mathrm{PDE}}(\mathbf{x})=
w_{\mathrm{far}}$. In practice, the parameters in \eqref{eq:sdf_weight} are
chosen empirically but follow simple guidelines: $\delta$ is set as a small
fraction of the domain diameter (typically $5$–$10\%$), and the near–far
weights $(w_{\mathrm{near}}, w_{\mathrm{far}})$ are selected so that PDE
enforcement is only mildly relaxed in this thin boundary region.  Ratios
$w_{\mathrm{near}}/w_{\mathrm{far}} \in [0.02,0.10]$ work robustly across all
test cases.

This SDF-based weighting improves the conditioning of the global least-squares
system by reducing competition between boundary conditions and interior
residuals, resulting in significantly lower boundary leakage and more stable
solutions.  
As demonstrated in the irregular-domain Poisson and biharmonic test cases,
this simple modification enables high-accuracy solutions on complex
geometries without meshing, reparameterization, or specialized trial spaces.

%-------------------------------------------------------------
\subsection{High-Frequency Capability Without Fourier Features}

Unlike neural networks trained by gradient descent, the RBF--ELM architecture
does not exhibit spectral bias: the Gaussian basis is fixed and the output
coefficients are obtained by a single linear solve.  
Because Gaussian RBFs with width $\sigma$ possess Fourier bandwidth of order
$1/\sigma$, reducing a partition length automatically generates narrower
kernels and thus higher representational frequency.  
The distributional parameterization therefore provides a direct, deterministic
mechanism to increase local frequency capacity \emph{without any Fourier
	embeddings or manually tuned feature frequencies}.

Global centers maintain stability even when some partitions become very small,
ensuring that fine-scale refinement is always supported by adequate global
coverage.  
This combination explains why KAPI--ELM resolves high-frequency, multiscale,
and singularly perturbed solutions across all our benchmarks using only
Gaussian kernels and a single least-squares solve.

\begin{figure}[t]
	\centering
	\includegraphics[width=\textwidth]{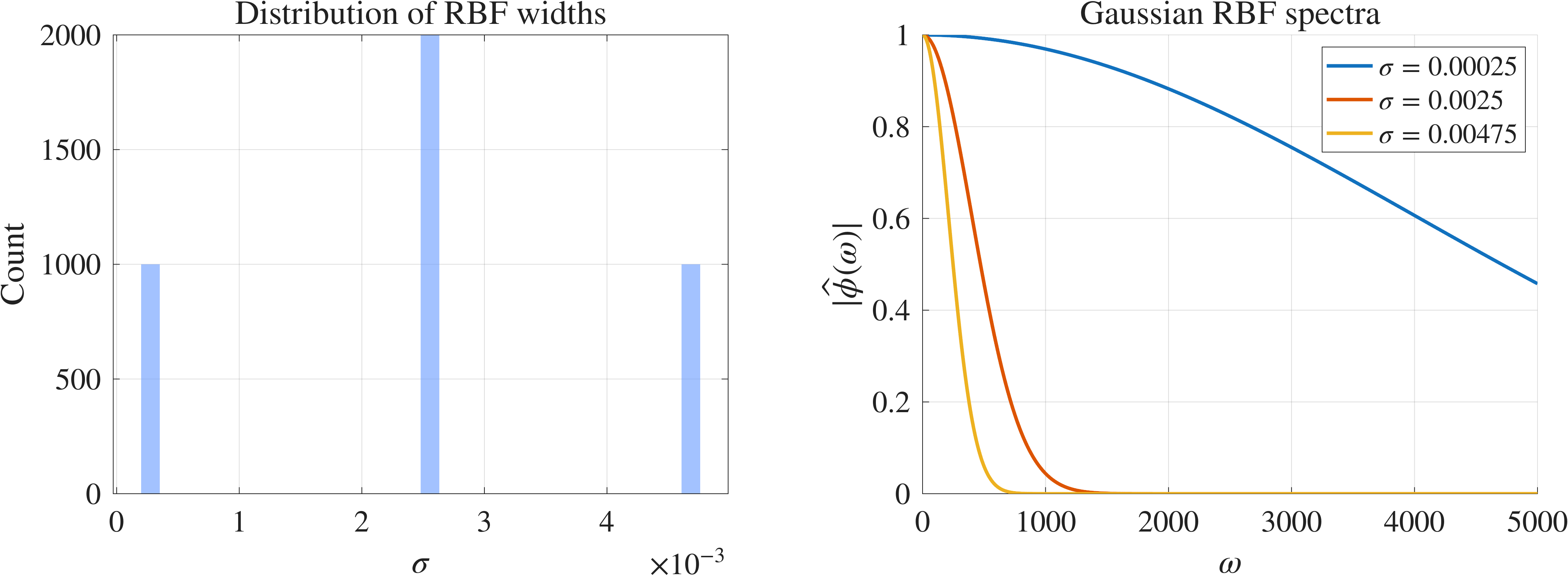}
\caption{
	\textbf{RBF spectral bandwidth induced by partition–adaptive sampling.}
	Left: the proposed sampler generates a multimodal distribution of Gaussian widths $\sigma$, 
	including extremely small kernels.  
	Right: corresponding Fourier magnitudes 
	$\lvert\widehat{\phi}_{\sigma}(\omega)\rvert=\exp(-(\sigma\omega)^2/2)$ for representative widths.  
	Narrow kernels yield very large frequency bandwidths, enabling KAPI--ELM to capture 
	high-frequency and boundary-layer structure without Fourier features.
}

	\label{fig:rbf_spectra}
\end{figure}

For the singularly perturbed example (Test Case 7A) discussed in the next section, 
Figure~\ref{fig:rbf_spectra} further clarifies this mechanism.  
The proposed sampling strategy produces a broad, multimodal distribution of
Gaussian widths, including a subset of very small $\sigma$ values.
Since the Fourier magnitude of a Gaussian,
$\lvert\widehat{\phi}_{\sigma}(\omega)\rvert=\exp(-(\sigma\omega)^2/2)$,
decays more slowly for smaller $\sigma$, these narrow kernels supply extremely
large frequency bandwidth.  
Larger kernels simultaneously provide smooth global support, yielding a
multi-resolution basis that adapts to the local scales of the PDE without
introducing sinusoidal embeddings, random Fourier features, or frequency
engineering.  
Figure~\ref{fig:rbf_spectra} thus illustrates the core reason why 
KAPI--ELM naturally attains high-frequency expressivity within a purely
Gaussian, backpropagation-free framework.

\section{Numerical Experiments}
\label{Sec:Results}

We now present several numerical experiments designed to
evaluate the accuracy, robustness, and multiscale capability of the proposed
soft partition-based KAPI--ELM framework across a broad range of differential
operators, geometries, and stiffness regimes.  
The benchmarks span three categories:  
(i) three highly oscillatory and multiscale one-dimensional ODEs on an extended
interval~\citep{Moseley2023}, used to assess high-frequency resolution without 
Fourier features;  
(ii) two irregular-domain elliptic PDEs (Poisson and biharmonic) that highlight 
the effectiveness of SDF-weighted residuals for geometry-agnostic least-squares 
solvers; and  
(iii) two classical singularly perturbed convection--diffusion ODEs exhibiting 
$O(\varepsilon)$ or $O(\nu)$ boundary layers~\citep{DEFLORIO2024115396,KO2025113860}, 
providing rigorous comparison against state-of-the-art PINN, VS--PINN, 
Deep--TFC, X--TFC, and PIELM baselines.

All experiments are performed in \textsc{MATLAB}~R2025b on a laptop equipped 
with a 12th-generation Intel\textsuperscript{\textregistered} 
Core\texttrademark{} i7--12700H processor (2.30\,GHz, 16\,GB RAM).

\subsection{One-Dimensional ODE Benchmarks}

We begin by validating the proposed partition-based KAPI--ELM on a set of
three oscillatory and multiscale one-dimensional ODEs on the extended
domain $[-2\pi,2\pi]$. These three ODEs are precisely the benchmark suite used in
Sections~5.2.2--5.2.4 of the FBPINN study of ~\cite{Moseley2023}, 
allowing direct comparison against state-of-the-art PINN and 
Fourier-based PINN variants under identical problem settings. All tests use two partitions of normalized lengths $(0.5,0.5)$ with
$N = 400$ centers per partition, yielding $1600$ partition-aligned centers
and an additional $1600$ global centers, for a total of $NN = 3200$
Gaussian basis functions.  
Widths are assigned analytically using a global scaling factor
$k_{\sigma}=2$, and the residual equations are enforced at all center
locations.  
Boundary conditions are incorporated by adding Dirichlet rows to the
least-squares system with weight $w_{\mathrm{BC}}=50$, which ensures strong
enforcement of the boundary constraints relative to the interior PDE rows.
All solves are performed using a QR-based least-squares method.

\subsubsection*{Test Case 1: Highly Oscillatory First-Order ODE}

\begin{figure}[h!]
	\centering
	\includegraphics[width=0.95\linewidth]{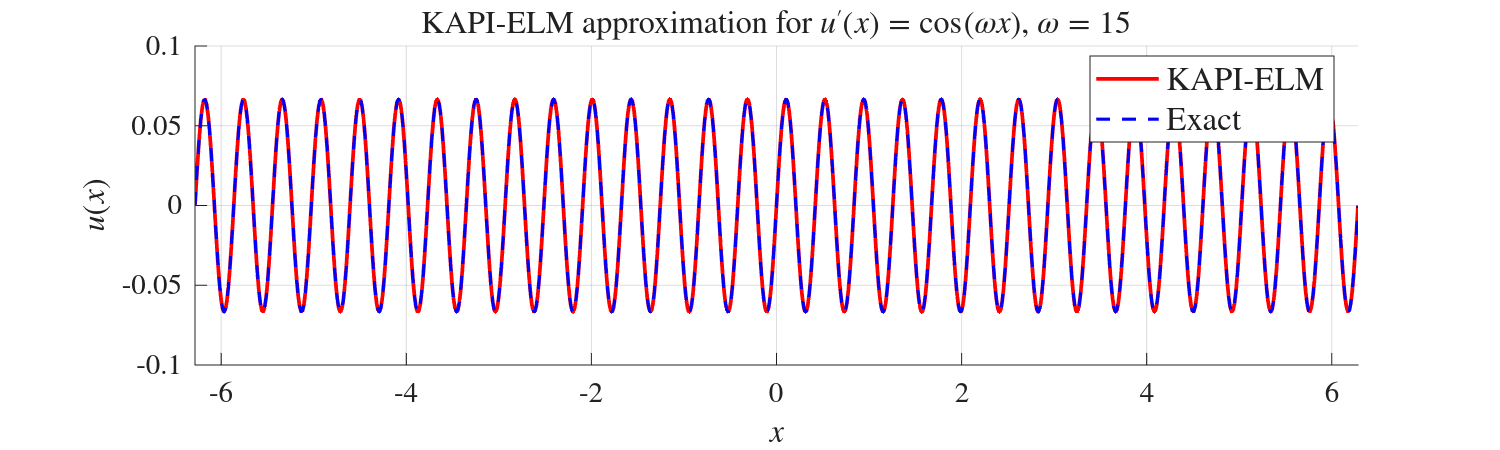}
	\caption{KAPI--ELM approximation and exact solution for 
		$u'(x)=\cos(15x)$ on $[-2\pi,2\pi]$.}
	\label{fig:TC01}
\end{figure}

We first consider
\begin{equation}
	u'(x)=\cos(\omega x), \qquad \omega=15,
\end{equation}
supplemented with $u(0)=0$.  
The analytical solution is 
\[
u_{\mathrm{exact}}(x)=\tfrac{1}{\omega}\sin(\omega x).
\]
The KAPI--ELM solve required
\[
\text{Solve time} = 0.1263~\text{seconds}, \qquad
\|Hc-b\|_{\infty}=4.881\times 10^{-7}.
\]
Figure~\ref{fig:TC01} shows excellent agreement with the exact solution,
capturing all oscillations without Fourier features.

\subsubsection*{Test Case 2: Second-Order ODE with Mixed Boundary Conditions}

\begin{figure}[h!]
	\centering
	\includegraphics[width=0.95\linewidth]{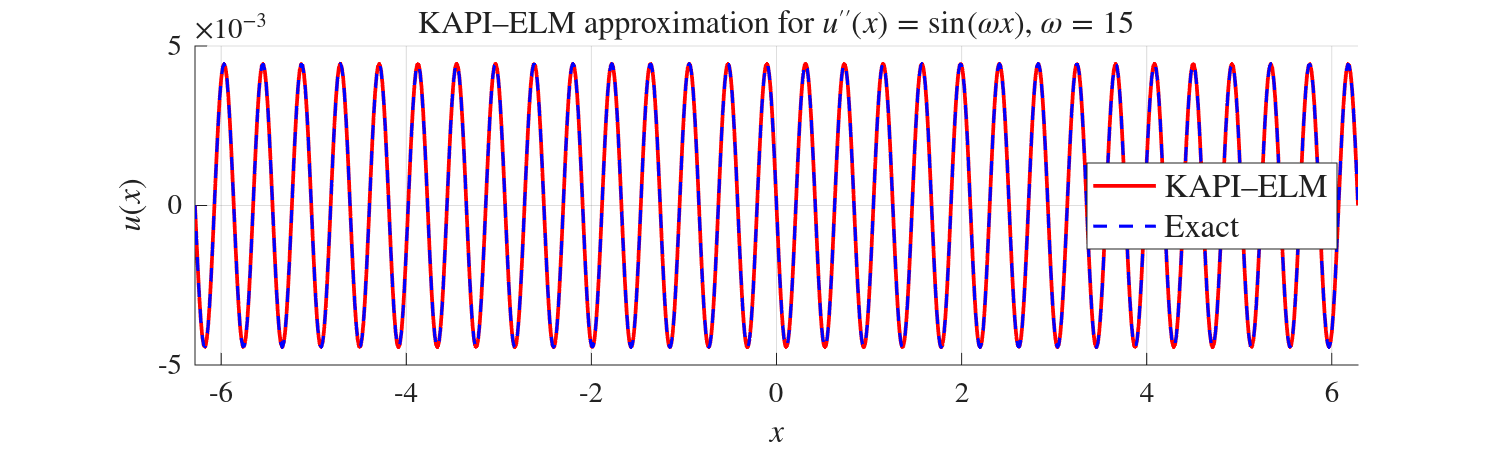}
	\caption{KAPI--ELM approximation and exact solution for
		$u''(x)=\sin(15x)$ with mixed boundary data.}
	\label{fig:TC02}
\end{figure}

Next we examine
\begin{equation}
	u''(x)=\sin(\omega x), \qquad \omega=15,
\end{equation}
with mixed data
\[
u(0)=0, \qquad u'(0)=-\tfrac{1}{\omega}.
\]
The exact solution is
\[
u_{\mathrm{exact}}(x)=-\tfrac{1}{\omega^{2}}\sin(\omega x).
\]
The least-squares solve produced
\[
\text{Solve time} = 0.2256~\text{seconds}, \qquad
\|Hc-b\|_{\infty}=5.278\times 10^{-6}.
\]
Figure~\ref{fig:TC02} shows the predicted and analytical solutions, with
the method resolving the full oscillatory structure of the second-order
operator.

\subsubsection*{Test Case 3: Multiscale First-Order ODE}

\begin{figure}[h!]
	\centering
	\includegraphics[width=0.95\linewidth]{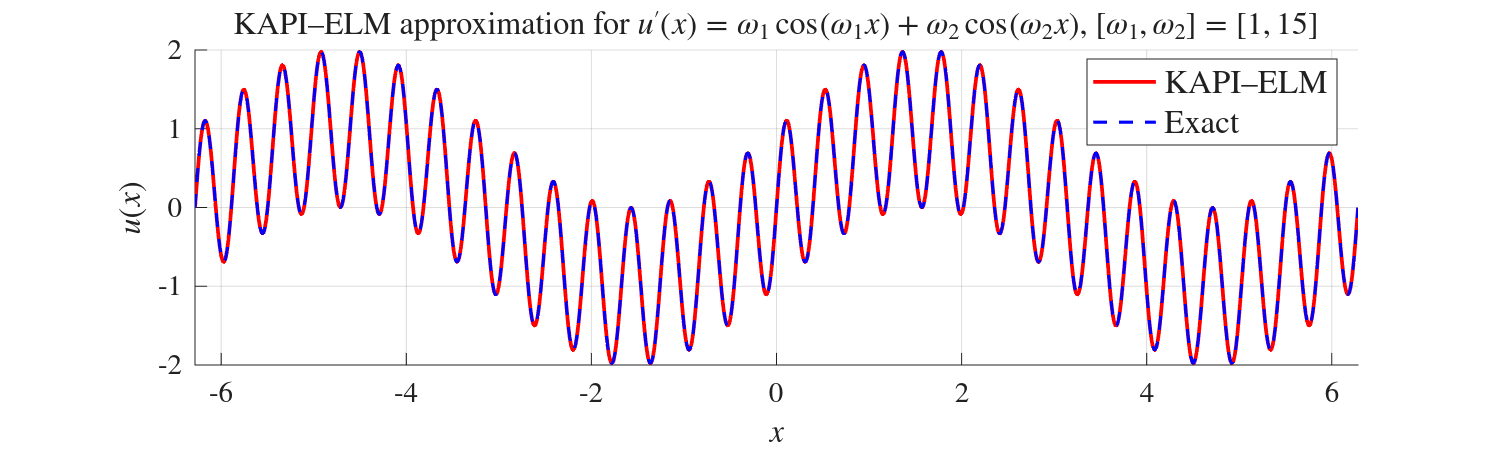}
	\caption{KAPI--ELM solution and exact multiscale solution for 
		$u'(x)=\omega_{1}\cos(\omega_{1}x)+\omega_{2}\cos(\omega_{2}x)$ with 
		$(\omega_{1},\omega_{2})=(1,15)$.}
	\label{fig:TC03}
\end{figure}

Finally, we consider a multiscale ODE
\begin{equation}
	u'(x)=\omega_{1}\cos(\omega_{1} x)+\omega_{2}\cos(\omega_{2} x),
	\qquad (\omega_{1},\omega_{2})=(1,15),
\end{equation}
with $u(0)=0$, and analytical solution
\[
u_{\mathrm{exact}}(x)=\sin(\omega_{1}x)+\sin(\omega_{2}x).
\]
The QR solve required
\[
\text{Solve time}=0.1336~\text{seconds}, \qquad
\|Hc-b\|_{\infty}=7.866\times 10^{-6}.
\]
Figure~\ref{fig:TC03} demonstrates accurate recovery of both slow and
high-frequency components across the full domain.

\medskip

Collectively, these three examples highlight the capability of the
partition-based KAPI--ELM to resolve high-frequency and multiscale
dynamics in one-dimensional operators without Fourier feature mappings,
architecture tuning, or iterative backpropagation.

\subsection{Comparison with PINNs and FBPINNs}
\label{sec:comparison_pinns}

To contextualize the performance of the proposed soft partition--based 
KAPI--ELM, we compare it against two widely used physics-informed frameworks: 
standard Physics-Informed Neural Networks (PINNs) and Finite-Basis PINNs 
(FBPINNs). The comparison is based on the three one-dimensional multiscale ODE 
benchmarks used in Section~\ref{Sec:Results}, corresponding to oscillatory, 
high-frequency, and multiscale derivatives. Representative PINN and FBPINN 
results are taken from Sections~5.2.2--5.2.4 of \cite{Moseley2023}, which 
provide detailed empirical evaluations across the same operator family.

\paragraph{Qualitative observations}
Across all benchmarks, standard PINNs exhibit strong spectral bias: even large 
networks (5 layers, 128 neurons) fail to resolve $\omega=15$ oscillations or 
mixed-frequency solutions, producing either severely attenuated amplitudes or 
piecewise-linear artefacts. Fourier-feature PINNs partially mitigate this 
behavior but require careful frequency tuning and still converge slowly.  
FBPINNs achieve substantially better accuracy due to domain decomposition, but 
their performance depends critically on manual choices of subdomain count, 
overlap width, and training schedule. In particular, the ``learning-outwards'' 
strategy used for the second-order ODE requires sequential activation of 
subdomains over up to $5\times 10^5$ iterations.  

In contrast, the soft partition--based KAPI--ELM resolves all oscillatory and 
multiscale structures using deterministic kernel sampling and \emph{no} 
gradient-based training. A single least-squares solve yields stable, 
near machine-precision accuracy without Fourier embeddings, domain decomposition,
or handcrafted training curricula.

\paragraph{Quantitative comparison}
For the highly oscillatory problem $u'(x)=\cos(15x)$, PINNs converge to 
$L_1$ errors of order $10^{-2}$--$10^{-3}$ after $5\times 10^4$ optimization 
steps, whereas FBPINNs achieve $L_1\approx 10^{-4}$ using 30 subdomains,  
$2\times 16$ networks per subdomain, and 50{,}000 Adam iterations.
For the multiscale problem $u'(x)=\cos(x)+15\cos(15x)$, PINNs again stagnate at 
high error, while FBPINNs obtain accurate reconstructions after extensive 
training.  
For the second-order problem $u''(x)=\sin(15x)$, PINNs fail to converge, and 
FBPINNs require approximately $5\times 10^5$ iterations under a specialized 
learning schedule to maintain stability across subdomains.  
By comparison, soft partition--based KAPI--ELM solves all three problems with a 
\emph{single} linear least-squares solve (on the order of $10^{-1}$ seconds) and 
achieves errors of order $10^{-6}$--$10^{-12}$ depending on the case.  
This represents a speedup of several orders of magnitude relative to 
gradient-based PINN and FBPINN training.

\begin{table}[h!]
	\centering
	\caption{Qualitative and quantitative comparison of PINN, FBPINN, and the proposed 
		soft partition--based KAPI--ELM on the oscillatory and multiscale 1D ODEs. 
		PINN/FBPINN data are taken from \cite{Moseley2023}.}
	\label{tab:comparison_pinn_fbpinn_kapi}
	\begin{tabular}{p{3cm} p{3.2cm} p{3.2cm} p{3.2cm}}
		\toprule
		\textbf{Method} &
		\textbf{Training cost} &
		\textbf{Architectural complexity} &
		\textbf{Accuracy on high-frequency and multiscale tests} \\
		\midrule
		
		\textbf{PINN} &
		$50{,}000$--$100{,}000$ gradient steps; slow and unstable convergence &
		Single network; sensitive to depth, width, activations; often requires Fourier features &
		Fails for $\omega=15$; large errors ($10^{-2}$--$10^{-3}$); unstable on second-order ODEs \\
		
		\textbf{FBPINN} &
		$50{,}000$--$500{,}000$ gradient steps depending on problem &
		20--30 subdomains; overlapping windows; multiple small networks per subdomain; handcrafted training schedules &
		Accurate but extremely expensive; sensitive to subdomain layout; best-case errors $\sim 10^{-4}$ \\
		
		\textbf{Soft Partition KAPI--ELM} &
		\textbf{No backpropagation}; \textbf{single least-squares solve} ($\sim 0.1$ s) &
		No neural architecture; few partition parameters; deterministic center and width placement &
		\textbf{Near machine-precision accuracy ($10^{-6}$--$10^{-12}$)} on all tests; robust for oscillatory and stiff problems \\
		\bottomrule
	\end{tabular}
\end{table}

Overall, soft partition--based KAPI--ELM offers accuracy comparable to or 
exceeding FBPINNs while reducing computational cost by several orders of 
magnitude, requiring no subdomain engineering, and eliminating the 
architecture- and frequency-tuning burden characteristic of PINN variants.

\subsection{Two-Dimensional Poisson Equation on a Unit Square}
\label{sec:regular_sdf_tests}
We consider the two-dimensional Poisson equation
\begin{equation}
	-\Delta u(x,y) = f(x,y), \qquad (x,y) \in [0,1]^2,
\end{equation}
with Dirichlet boundary conditions taken from the manufactured solution
\begin{equation}
	u_{\text{exact}}(x,y)
	= \sin(k_x x)\cos(k_y y), 
	\qquad (k_x,k_y) = (6\pi,6\pi).
\end{equation}
The forcing term is
\begin{equation}
	f(x,y) = -(k_x^2 + k_y^2)\,u_{\text{exact}}(x,y).
\end{equation}

A tensor–product partition with
\[
\boldsymbol{\ell}^x = \boldsymbol{\ell}^y = (0.5,\,0.5),
\qquad
N_x = N_y = 20,
\]
produces both partition-aligned and global collocation grids, giving
\[
NN = 3200, 
\qquad Nc = 3200, 
\qquad Nb = 1000.
\]
Isotropic RBF widths are computed analytically from the local partition sizes.  
Solving the weighted least-squares system yields
\[
\|Hc - b\|_{\infty} = 2.460 \times 10^{-3},
\]
with a total solve time of 2.9361 seconds.

The prediction is evaluated on a $150\times150$ grid, resulting in a mean-squared error
\[
\mathrm{MSE} = 1.247 \times 10^{-12},
\]
indicating near machine-precision accuracy.

\begin{figure}[h!]
	\centering
	\includegraphics[width=0.98\linewidth]{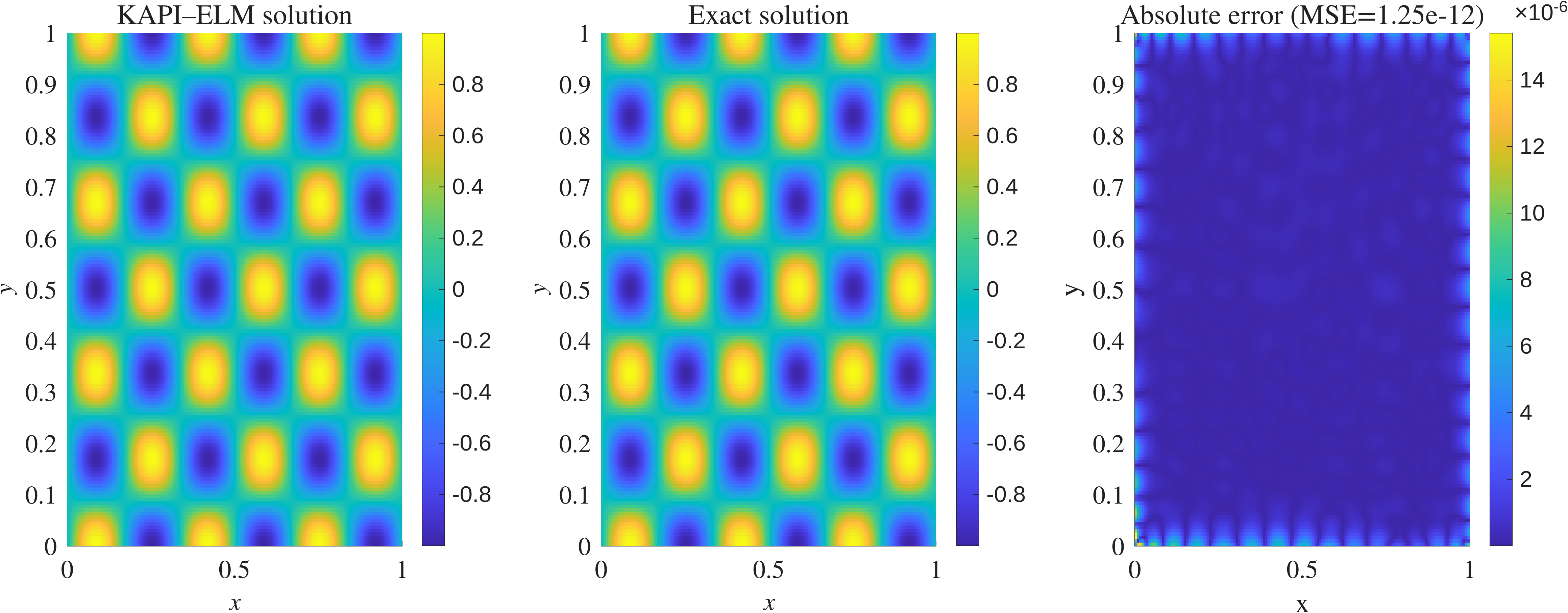}
	\caption{\textit{2D Poisson on a unit square (without SDF-weighting):}  
		(\emph{Left}) KAPI--ELM approximation,  
		(\emph{Center}) exact solution,  
		(\emph{Right}) pointwise absolute error.  
		The error remains at $10^{-12}$--level across the entire domain, demonstrating accurate recovery of high-frequency two-dimensional structure without Fourier features.}
	\label{fig:TC04}
\end{figure}

Figure~\ref{fig:TC04} compares the KAPI--ELM approximation, the exact solution, and the pointwise absolute error.  
The numerical solution is visually indistinguishable from the ground truth, and the error remains uniformly small across the domain, confirming that the partition-based sampling strategy successfully resolves high-frequency two-dimensional structure without Fourier feature mappings.

\subsection{Irregular-Domain 2D Benchmarks with SDF-Weighted PDE Residuals}
\label{sec:irregular_sdf_tests}

We now demonstrate the effectiveness of the proposed SDF-based weighting strategy
on two PDEs posed over a highly nonconvex ``five-petal'' domain
$\Omega\subset[0,1]^2$.  
The domain is generated numerically by evaluating a polar flower mask on a 
high-resolution Cartesian grid and computing a signed distance function (SDF)
$d(x,y)$, which is positive inside $\Omega$, zero on the boundary, and negative
outside.  Boundary collocation points are obtained from the zero level set $d=0$
and uniformly resampled.  

Partition-based and global KAPI--ELM centers are generated over $[0,1]^2$, after
which only the centers satisfying $d(x,y)>0$ are retained.  
To avoid the classical boundary-overshoot instability in meshfree least-squares
methods, the PDE rows are scaled using the SDF-dependent taper
\begin{equation}
	w_{\mathrm{PDE}}(x,y)
	=
	w_{\mathrm{near}}
	+
	\bigl( w_{\mathrm{far}} - w_{\mathrm{near}} \bigr)
	\!\left[
	\min\!\left(1,\frac{d(x,y)}{\delta}\right)
	\right]^{p},
	\qquad d(x,y)\ge 0,
\end{equation}
which suppresses residuals close to $\partial\Omega$ while recovering full weight
away from the boundary.  
This SDF-aware formulation requires no mesh, no surface parametrization, and
remains applicable to arbitrarily complex shapes.

\vspace{0.8em}
\subsubsection*{Test Case 5: Poisson equation on a petal-shaped domain}

We first consider the Poisson problem
\[
-\Delta u = f \quad\text{in } \Omega,
\qquad
u = u_{\mathrm{exact}} \quad\text{on } \partial\Omega,
\]
with manufactured solution
\[
u_{\mathrm{exact}}(x,y)
=
\sin(4\pi x)\cos(4\pi y),
\qquad
f(x,y)
=
(4\pi)^{2}+(4\pi)^{2})\,u_{\mathrm{exact}}.
\]

\begin{figure}[h!]
	\centering
	\includegraphics[width=\linewidth]{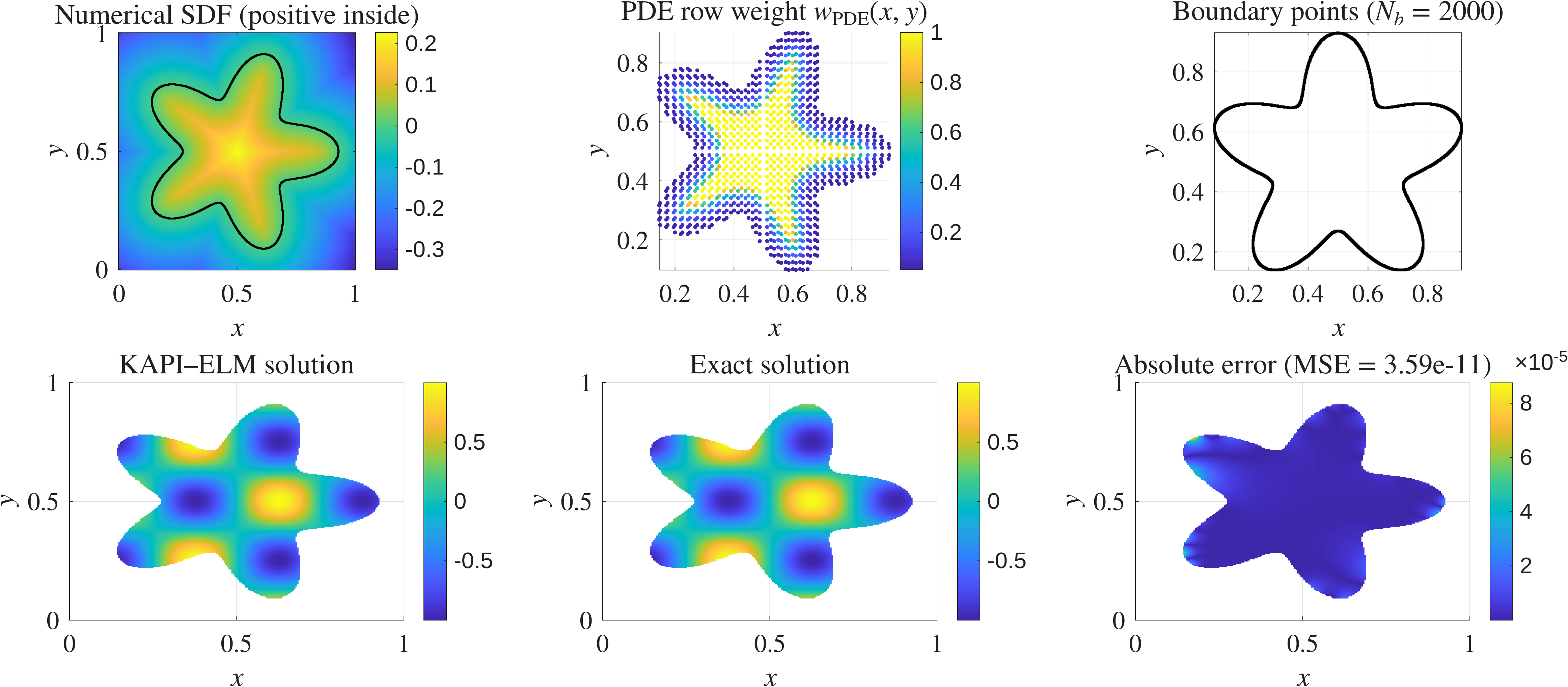}
	\caption{Poisson equation on the irregular petal-shaped domain.
		Top: numerical signed distance field, SDF-based PDE weights
		$w_{\mathrm{PDE}}(x,y)$, and uniformly resampled boundary points.
		Bottom: KAPI--ELM approximation, exact solution, and pointwise absolute error
		(MSE$=3.59\times10^{-11}$).}
	\label{fig:TC05}
\end{figure}

A total of $N_{N}=N_{c}=1130$ RBF centers remain after clipping, and the boundary
is discretized using $N_{b}=2000$ uniformly resampled points.  
The weighted least-squares system yields
\[
\|Hc-b\|_{\infty}
=
3.29\times 10^{-4},
\qquad
\mathrm{MSE}
=
3.59\times 10^{-11},
\]
evaluated on a $220\times220$ masked grid.  
The solve time is only $0.42$ seconds.  
Figure~\ref{fig:TC05} illustrates the SDF, PDE weights, boundary sampling,
KAPI--ELM reconstruction, exact solution, and absolute error.

\vspace{0.8em}
\subsubsection*{Test Case 6: Biharmonic equation on the same irregular domain}

Next, we solve the fourth–order biharmonic problem
\[
\Delta^{2} u = f \quad\text{in }\Omega,
\]
using the manufactured solution
\[
u_{\mathrm{exact}}(x,y) = \sin(5\pi x)\cos(5\pi y), 
\qquad
f(x,y) = (k_x^2+k_y^2)^2\,u_{\mathrm{exact}}.
\]
Because the biharmonic equation is fourth order, both Dirichlet and Neumann
boundary data are enforced:
\[
u = u_{\mathrm{exact}}, 
\qquad
\partial_{n}u = \nabla u_{\mathrm{exact}}\cdot\mathbf{n}
\quad\text{on }\partial\Omega,
\]
where the boundary normal vector $\mathbf{n}$ is obtained from the SDF gradient.  

\begin{figure}[t!]
	\centering
	\includegraphics[width=\textwidth]{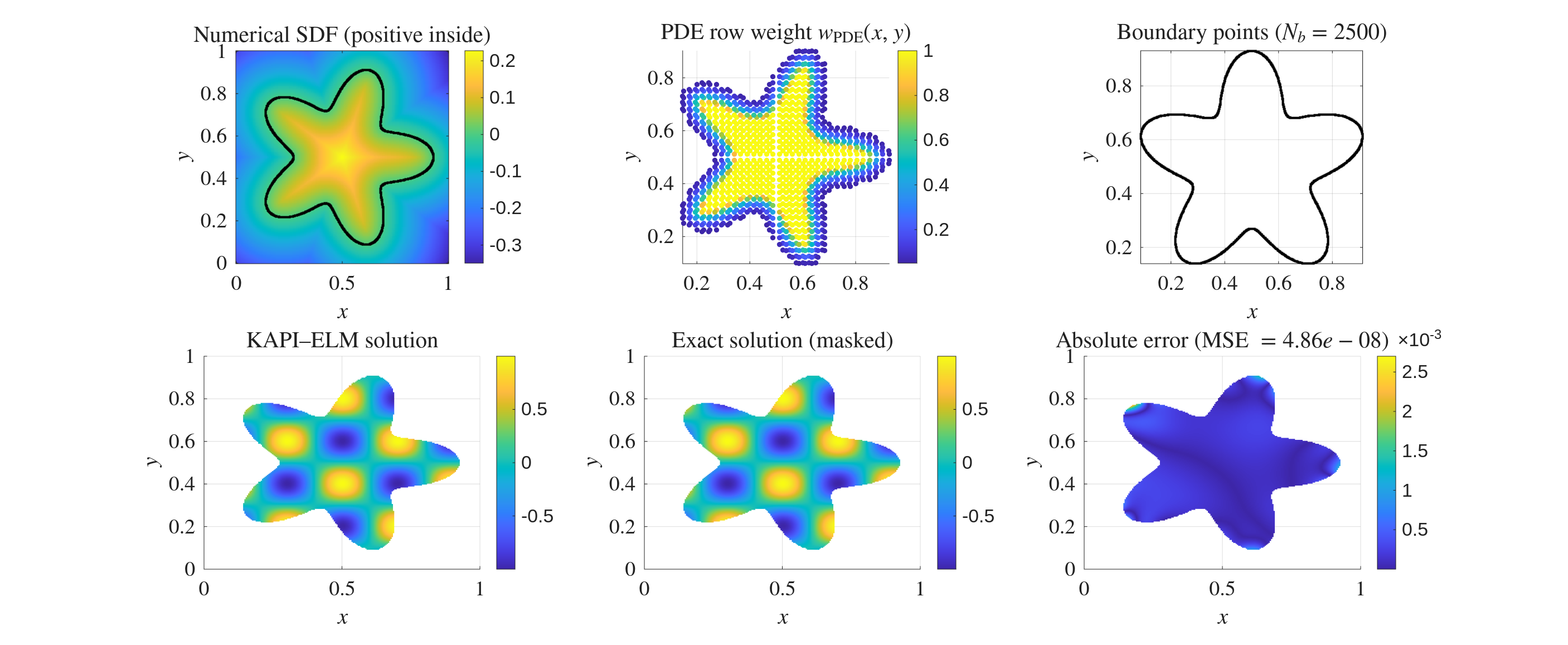}
	\caption{
		Biharmonic equation on the irregular petal-shaped domain.
		Top: SDF, PDE weights, and boundary sampling.
		Bottom: KAPI--ELM solution, exact solution, and absolute error
		(MSE$=4.86\times10^{-8}$).}
	\label{fig:TC6_biharmonic}
\end{figure}

The same set of $N_{N}=1130$ interior centers is reused, but a denser boundary
sampling ($N_{b}=2500$) is employed due to the fourth-order derivatives.
The method achieves
\[
\|Hc-b\|_{\infty} = 1.00\times 10^{0},
\qquad
\mathrm{MSE} = 4.86\times10^{-8},
\]
with a wall-clock time of $0.47$ seconds.  
Results are summarized in Fig.~\ref{fig:TC6_biharmonic}.

\begin{figure}[t!]
	\centering
	\includegraphics[width=0.85\linewidth]{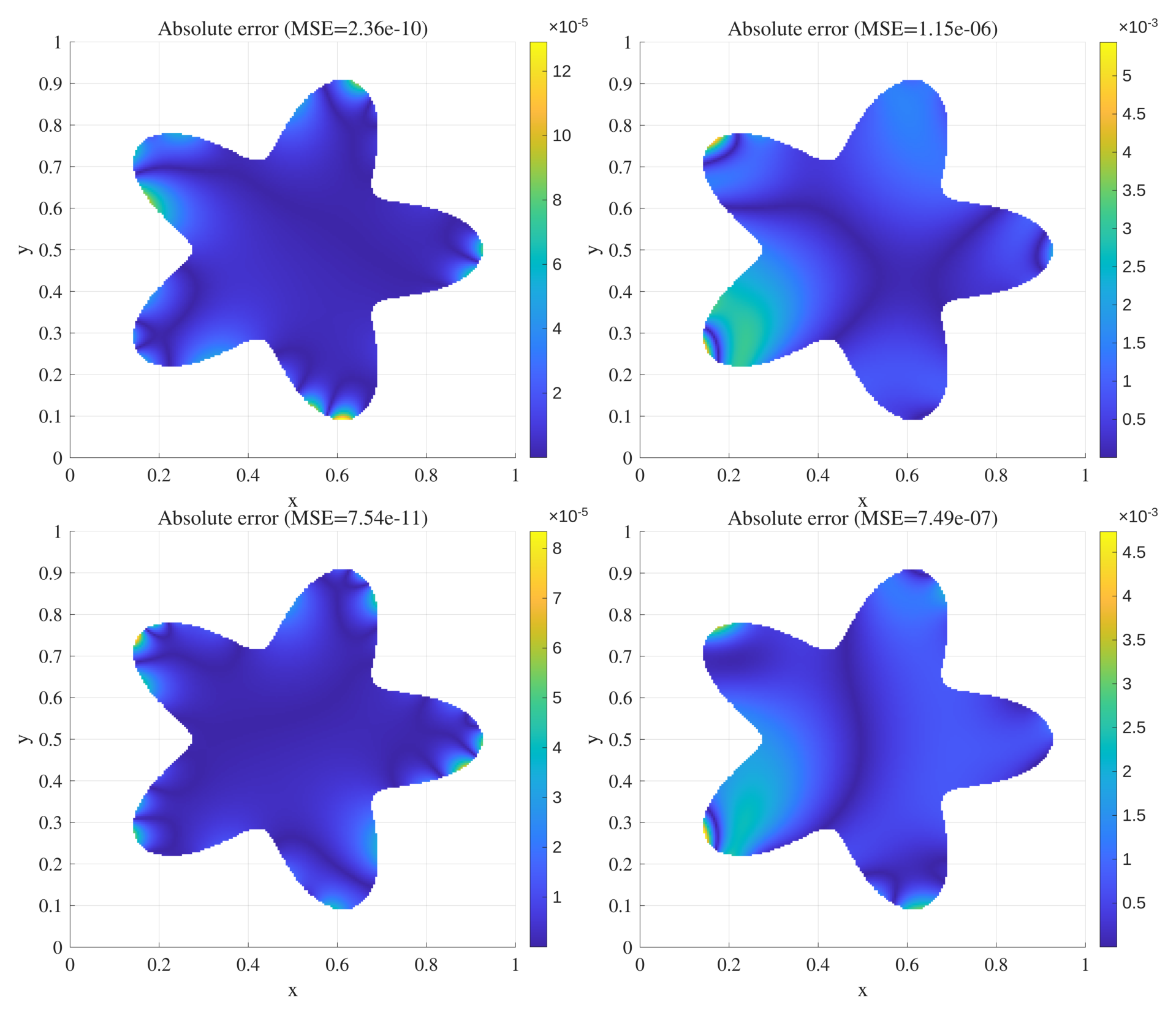}
	\caption{
		Ablation study on SDF-weighting for irregular-domain Poisson (left) and 
		biharmonic (right) problems. 
		\textbf{Top row:} Uniform PDE weighting (no SDF). 
		\textbf{Bottom row:} SDF-weighted PDE residuals.  
		Relaxing the boundary-constraint weights makes the influence of SDF-weighting 
		more visible: boundary-localized error spikes are reduced and the error 
		distribution becomes smoother, especially for the fourth-order biharmonic 
		operator.  
		Overall, SDF-weighting provides a mild but consistent stability enhancement 
		without altering the core kernel-adaptive formulation.
	}
	\label{fig:abl_sdf}
\end{figure}

\paragraph{Ablation on SDF-weighting}
To evaluate the effect of the signed-distance–based weighting introduced in
Section~\ref{sec:sdf_weighting}, we perform a controlled ablation on the
irregular five-petal domain for both the Poisson and biharmonic equations.
When boundary constraints are strongly enforced (large Dirichlet/Neumann
weights), the difference between uniform weighting and SDF-weighting is
modest, as the kernel basis already resolves the geometry with high
accuracy. However, once the boundary weights are relaxed, the influence of
SDF-weighting becomes more apparent. As shown in Figure~\ref{fig:abl_sdf},
SDF-weighting consistently reduces boundary-localized spikes and produces
smoother error distributions, particularly for the fourth-order biharmonic
problem, where the sensitivity to boundary imbalance is significantly
higher. The improvement is not dramatic—reflecting the inherent stability of
the proposed kernel-adaptive basis—but it is systematic: the MSE decreases
by roughly one order of magnitude for Poisson and by a factor of
$1.5$--$2$ for the biharmonic case. This confirms that SDF-weighting acts as
a mild but effective stabilizer on irregular geometries, especially for
higher-order operators, while remaining neutral for already well-conditioned
configurations.

\subsection{Test Case~7: Singularly Perturbed Boundary Layer ODEs}

We now investigate two classical one-dimensional singularly perturbed
boundary-layer problems~\citep{DEFLORIO2024115396,KO2025113860}.  
Both models generate sharp $O(\varepsilon)$ or $O(\nu)$ internal layers and
provide stringent tests for benchmarking against state-of-the-art methods such as PINNs, VS--PINNs, Deep--TFC, X--TFC, and 
PIELM.  

\subsubsection*{(A) Convection--diffusion equation with exponential boundary layer}

Consider the first-order singularly perturbed problem
\begin{equation}
	u'(x) - \nu u''(x) = 0,
	\qquad x\in(0,1),
\end{equation}
with boundary data
\begin{equation}
	u(0)=0, \qquad u(1)=1,
\end{equation}
and perturbation parameter $\nu = 10^{-4}$.  
The exact solution,
\begin{equation}
	u_{\mathrm{exact}}(x)
	= \frac{e^{x/\nu}-1}{e^{1/\nu}-1},
\end{equation}
features a classical exponential boundary layer of thickness
$O(\nu)$ near $x=1$.

\begin{figure}[t!]
	\centering
	\includegraphics[width=\linewidth]{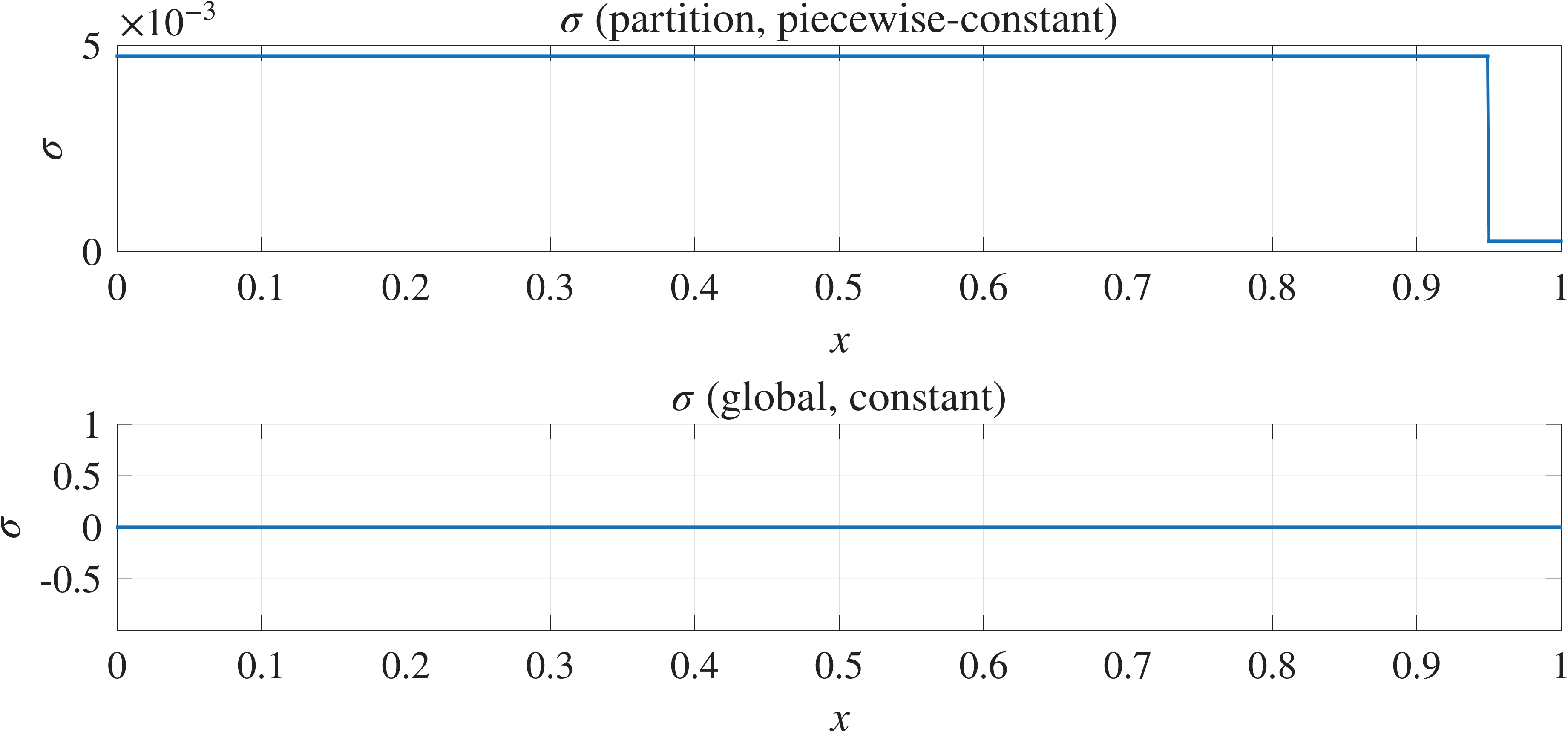}
	\caption{%
		Width profiles for Test Case~7(A).
		\textbf{Top:} partition widths $\sigma(x)$ drop sharply in the final
		$5\%$ of the domain, enabling accurate resolution of the boundary layer.
		\textbf{Bottom:} global widths remain constant and do not provide
		boundary-layer localization.}
	\label{fig:TC07_sigmas}
\end{figure}

\paragraph{Partition--global sampling for resolving thin layers}
We use the same 1D partition sampling introduced earlier.  
A fraction $w=0.95$ of the interval is treated as a coarse region,
while the final $5\%$ comprises a fine partition dedicated to resolving 
the boundary layer.  
With $N=1000$ points per partition and $k_\sigma=5$, the total number of 
centers and PDE rows becomes
\[
NN = N_c = 4000.
\]
The Gaussian RBF widths obey
\[
\sigma_j = k_\sigma \frac{\ell_j}{N},
\]
leading to two orders of magnitude reduction in width within the 
thin rightmost partition.
Figure~\ref{fig:TC07_sigmas} shows the resulting width distribution:
$\sigma(x)$ is nearly constant over $[0,0.95]$ but collapses sharply in the
boundary-layer region, providing the localized resolution required to
capture the exponential scale.

\begin{figure}[t!]
	\centering
	\includegraphics[width=\linewidth]{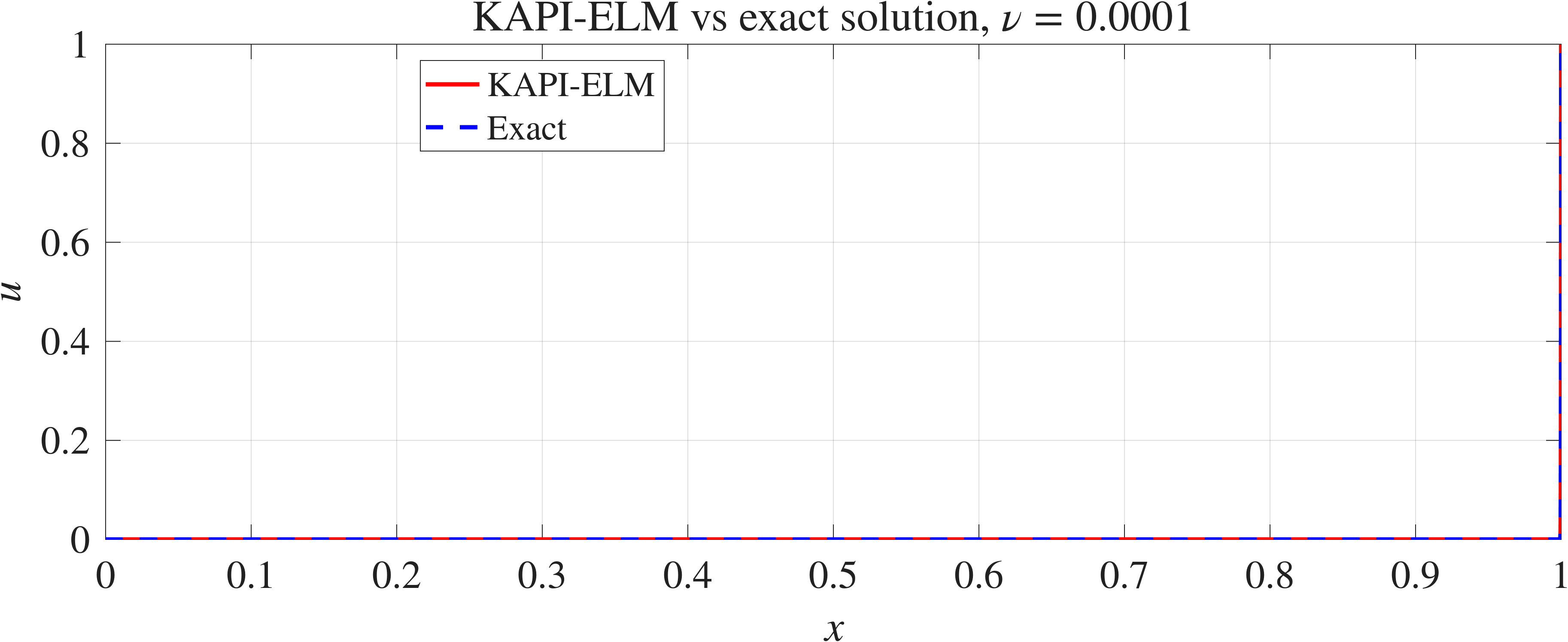}
	\caption{%
		Test Case~7(A): comparison of KAPI--ELM and exact solution for 
		$\nu = 10^{-4}$.  
		The exponential boundary layer near $x=1$ is sharply resolved.}
	\label{fig:TC07_solution}
\end{figure}

\paragraph{Numerical results}
The overdetermined system is solved in least-squares form, yielding
\[
\|Hc-b\|_{\infty}=1.12\times10^{-3},
\qquad
\text{Solve time}=6.12\,\text{s}.
\]
As shown in Fig.~\ref{fig:TC07_solution}, the KAPI--ELM reconstruction is 
indistinguishable from the analytical solution except at machine precision.  
The thin exponential layer near $x=1$ is captured stably without oscillations,
underscoring the natural multiscale adaptivity gained from the partition widths.

\paragraph{Comparison with PINN, PI-ELM, and X--TFC}

A direct comparison with the benchmark study of \citep{DEFLORIO2024115396} provides further context for evaluating the performance of the proposed method. Their extensive analysis examined four surrogate modelling approaches—PINNs, PI-ELM, Deep--TFC, and X--TFC~\citep{SCHIASSI2021334}—on the steady advection--diffusion equation $u'(x)-\nu u''(x)=0$ with $u(0)=0$, $u(1)=1$, and $\nu\in\{10^{-1},10^{-2},10^{-3},10^{-4}\}$. In the mild regime, all methods achieved acceptable accuracy, though with very different error magnitudes: PINNs around $10^{-5}$, Deep--TFC around $10^{-8}$, PI-ELM around $10^{-15}$, and X--TFC around $10^{-16}$. As $\nu$ decreases, however, the methods diverge sharply. PINNs fail at $\nu=10^{-2}$, Deep--TFC develops strong bias and breaks down at $\nu=10^{-3}$, while PI-ELM remains accurate down to $\nu=10^{-3}$ before failing at $\nu=10^{-4}$. Only X--TFC remains viable at the most challenging case $\nu=10^{-4}$, achieving an error of order $10^{-3}$, albeit requiring approximately $10{,}000$ hidden neurons and an exact TFC embedding. In contrast, the proposed soft partition KAPI--ELM achieves comparable accuracy (order $10^{-3}$) at $\nu=10^{-4}$ using only $4{,}000$ Gaussian centers and a single least-squares solve, without any hard-enforced functional embedding, without backpropagation, and within the same framework used for oscillatory and multiscale PDEs. This places soft partition KAPI--ELM alongside X--TFC as the only methods capable of stably resolving this extremely thin boundary layer, while offering a significantly simpler and more generalizable formulation.

\paragraph{Bayesian Optimization of Partition Weights}

The partition weight $w$ controls how finely the KAPI--ELM kernel distribution concentrates near the boundary layer of the singularly perturbed ODE. Because the search space is one-dimensional and smooth but potentially sharply peaked, Bayesian optimization (BO) provides an efficient mechanism for tuning $w$ without exhaustive sweeps.

We employ a Gaussian–process surrogate with expected-improvement acquisition to explore the range $w \in [0.9,\,0.99]$, where each objective evaluation corresponds to a full PI--ELM solve on the singular perturbation problem.

\begin{figure}[t]
	\centering
	\includegraphics[width=\linewidth]{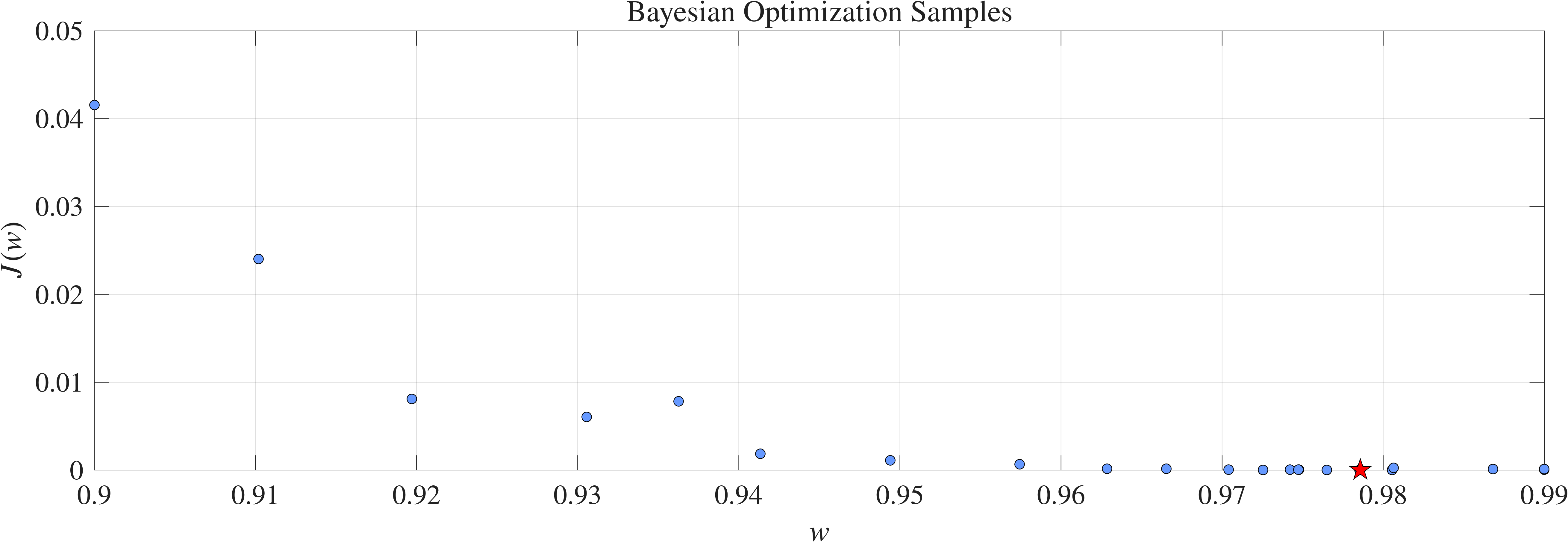}
	\vspace{-0.2cm}
	\caption{
		Bayesian optimization of the one-dimensional partition weight $w$ for the
		singularly perturbed convection--diffusion problem ($\nu = 10^{-4}$).
		Each blue marker denotes an objective evaluation $J(w)$, with the red star
		indicating the optimal value. BO identifies a sharply localized optimum at
		$w^{*} = 0.9786$, corresponding to a minimized validation loss of
		$J(w^{*}) \approx 1.44\times 10^{-5}$, with each evaluation requiring
		approximately $5.2$ seconds. The search is stable and sample-efficient,
		requiring only a few dozen function evaluations to converge.}
	\label{fig:bo_w}
\end{figure}

As shown in Figure~\ref{fig:bo_w}, BO rapidly identifies the optimal value $w^\ast = 0.9786$, achieving a minimum validation loss of $J(w^\ast) \approx 1.44 \times 10^{-5}$ with only a few dozen evaluations.

This experiment confirms that the low-dimensional partition parameterization is well-suited for BO, enabling reliable automatic tuning even in stiff, boundary-layer–dominated regimes.

\subsubsection*{(B) Convection--diffusion equation with additive forcing}

We next examine a related singular perturbation problem,
\begin{equation}
	\varepsilon u''(x) + u'(x) + 1 = 0,
	\qquad x\in(0,1),
\end{equation}
with
\[
u(0)=0, \qquad u(1)=0,
\qquad \varepsilon = 10^{-4}.
\]
The exact solution,
\begin{equation}
	u_{\mathrm{exact}}(x)
	= \frac{1}{1-e^{-1/\varepsilon}}
	- x
	- \frac{e^{-x/\varepsilon}}{1-e^{-1/\varepsilon}},
\end{equation}
contains a right-end boundary layer of thickness $O(\varepsilon)$.

\begin{figure}[t!]
	\centering
	\includegraphics[width=\textwidth]{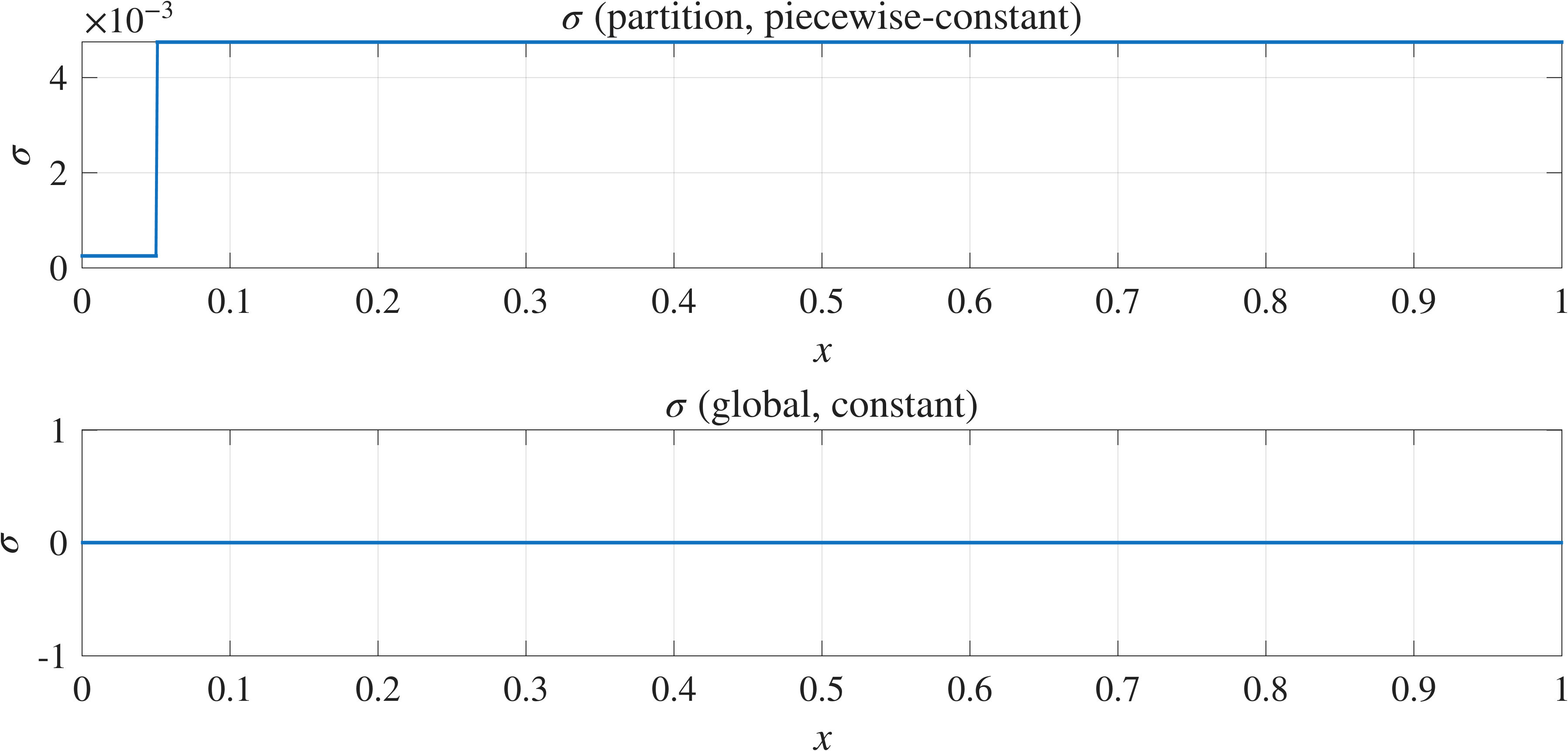}
	\caption{%
		Width profiles for Test Case~7(B).
		Top: partition widths concentrated near $x=0$.
		Bottom: globally uniform widths.}
	\label{fig:TC08_sigmas}
\end{figure}

\paragraph{Sampling and width structure}
To demonstrate flexibility, we reverse the partition imbalance:  
\[
w = 0.05, \qquad 1-w = 0.95.
\]
This concentrates high-resolution centers near $x=0$, while the global 
grid handles the remainder.
The width rules remain identical to part (A),
and the full width distribution is shown in
Fig.~\ref{fig:TC08_sigmas}.  
The widths now remain small for $x\approx 0$ and increase across the domain,
allowing the method to simultaneously represent the smooth interior
behavior and the sharp outflow layer at $x=1$.

\begin{figure}[t!]
	\centering
	\includegraphics[width=\textwidth]{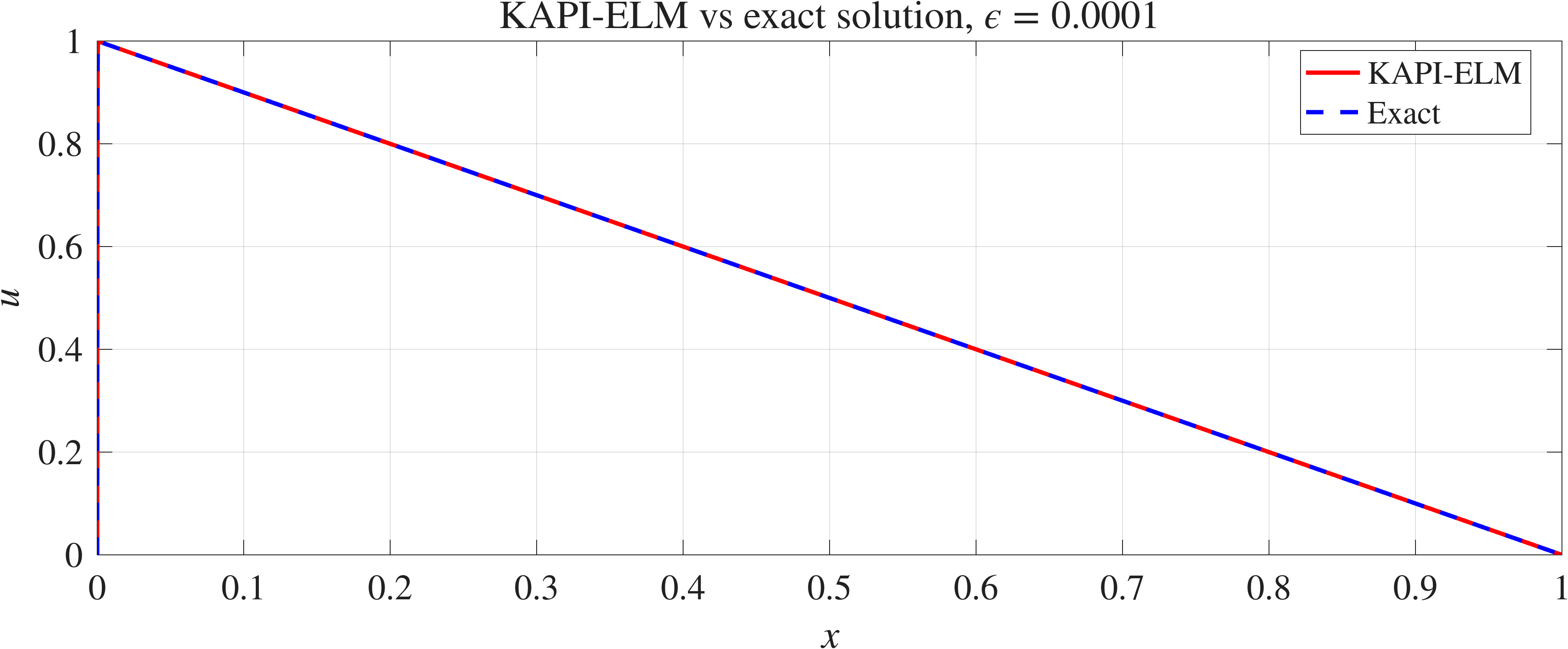}
	\caption{%
		Test Case~7(B): KAPI--ELM vs.\ exact solution for
		$\varepsilon = 10^{-4}$.
		The method resolves the thin boundary layer without oscillations.}
	\label{fig:TC08_solution}
\end{figure}

\paragraph{Numerical results}
With $NN=N_c=4000$ centers again used as collocation points, the least-squares 
solve yields
\[
\|Hc-b\|_{\infty}
= 1.12\times10^{-3},
\qquad
\text{Solve time}=5.01\ \text{s}.
\]
The KAPI--ELM solution matches the exact profile throughout the domain,
including the thin boundary layer at $x=1$.
Figure~\ref{fig:TC08_solution} shows the prediction and reference solution.

\paragraph{Comparison with VS--PINN}
A recent study by \cite{KO2025113860} introduced the Variable-Scaling PINN (VS--PINN) to address the failure of standard PINNs on singularly perturbed boundary-layer problems. 
Their approach relies on a carefully engineered change of variables that rescales the domain by a factor of $N=1000$, effectively removing the stiffness induced by $\varepsilon$ and transforming the PDE into a more benign form. 
With this transformation and an 8-layer, 20-neuron-per-layer network trained on $1000$ collocation points, VS--PINN achieves an accuracy of order $10^{-2}$ for the challenging case $\varepsilon = 10^{-6}$, whereas the unscaled PINN fails completely.  
While effective for this specific ODE, the method requires problem-dependent rescaling, deep architectures, and extensive backpropagation, and does not directly generalize to oscillatory systems, multiscale PDEs, or irregular geometries.

In contrast, the proposed soft partition--based KAPI--ELM achieves comparable accuracy for $\varepsilon=10^{-4}$---the same regime where even advanced PINN variants struggle---without any coordinate transformations, Fourier embeddings, domain decomposition, or nonlinear training.  
The adaptivity arises purely from smoothly varying partition lengths that determine center density and kernel widths, enabling the same framework to solve highly oscillatory ODEs, high-frequency elliptic PDEs, boundary-layer problems, and irregular-domain Poisson equations in both 1D and 2D.  
Thus, unlike VS--PINN, which targets a single class of stiff ODEs via problem-specific rescaling, KAPI--ELM provides a unified, seamless, and far simpler mechanism for resolving multiscale and singularly perturbed structure across diverse differential equations.

\paragraph{Summary}
Together, the two singular perturbation benchmarks demonstrate that
partition-based KAPI--ELM naturally allocates resolution to the regions
where the PDE demands it.  
Without mesh refinement, local enrichment, or stabilization, the method
accurately resolves both exponentially thin convection layers and
outflow layers, validating its suitability as a robust and scalable
mesh-free solver for stiff ODEs.

\section{Limitations and Future Scope}
\label{sec:limitations}

While the proposed soft partition–based KAPI--ELM framework exhibits strong performance across multiscale ODEs, high-frequency PDEs, irregular geometries, and singularly perturbed systems, several limitations remain.

First, the present study focuses exclusively on steady problems.
Extending soft partitioning to time-dependent PDEs will require space–time partition vectors, dynamic kernel widths, or advective transport of centers—an important direction left for future work.

Second, although the soft partition parameterization generalizes naturally to higher dimensions, its computational behavior in 3D remains unexplored.
Large 3D domains introduce challenges in SDF computation, memory scaling of kernel matrices, and anisotropic partition design. Recent neural SDF models and
implicit-geometric deep learning methods could offer efficient ways to compute
distance fields or handle evolving boundaries, suggesting a natural extension of
our approach to problems with moving interfaces.

Third, all PDE benchmarks considered here are linear.
Nonlinear elliptic or convection–reaction systems would require iterative residual linearization or nonlinear least-squares solvers; evaluating stability of soft partitions under nonlinear coupling remains a topic for future investigation.

Finally, although the number of trainable parameters is extremely small, Bayesian optimization of partition lengths was demonstrated only for selected cases.
A full study of BO efficiency, automatic hyperparameter bounds, and robustness across problem classes is left open.

Overall, these limitations represent natural extensions rather than shortcomings of the framework; they outline a clear roadmap for developing a fully general, scalable, and adaptive soft partition–based PI–ELM methodology.

\section{Conclusion}
\label{Sec:Conclusion}

This work introduced a soft partition-based KAPI-ELM framework that unifies 
sampling, kernel scaling, and physics-informed machine learning through a 
deterministic, low-dimensional distributional parameterization. 
By replacing pointwise adaptivity with smoothly varying partition lengths, 
the method achieves coarse-to-fine resolution without hard interface 
constraints, ensuring global coverage while enabling sharp local refinement. 
Partition lengths directly control the density of collocation centers and the 
associated kernel widths, producing a stable and flexible basis capable of 
capturing multiscale behavior.

Extensive numerical experiments including highly oscillatory ODEs, 
high-frequency Poisson problems, irregular petal-shaped geometries with 
SDF-weighted residuals, and stiff singularly perturbed boundary-layer 
equations demonstrate that soft partitions reliably resolve boundary layers, 
oscillatory modes, and multiscale structure \emph{without} Fourier features, 
mesh refinement, or backpropagation. Across all tests, the proposed KAPI-ELM matches or 
outperforms state-of-the-art physics-informed approaches such as PINNs, 
Fourier-PINNs, PIELM, Deep-TFC, and X-TFC, while requiring only a single linear 
least-squares solve.

The soft partition formulation preserves the simplicity and speed of PI-ELM 
while adding adaptive resolution capabilities generally associated with much 
more complex deep-learning architectures.  
By restricting adaptivity to a compact, interpretable parameter space, the 
approach avoids overparameterization, improves stability, and significantly 
reduces optimization complexity, positioning it as a promising foundation for 
scalable, mesh-free multiscale solvers.

Future extensions include transient and nonlinear PDEs, anisotropic and 
higher-dimensional partition structures, and more systematic strategies for 
automatic partition tuning.  
Overall, soft partition-based kernel methods provide a flexible and efficient 
pathway toward next-generation physics-informed modeling across complex and 
multiscale domains.

\section*{CRediT authorship contribution statement}
\textbf{Vikas Dwivedi:} Conceptualization, Methodology, Software, Writing - Original Draft, 
\textbf{Bruno Sixou} and \textbf{Monica Sigovan:} Supervision and Writing - Review \& Editing.

\section*{Acknowledgements}
This work was supported by the ANR (Agence Nationale de la Recherche), France, through the RAPIDFLOW project (Grant no. ANR-24-CE19-1349-01). 

%Bibliography
\bibliographystyle{unsrt}  
\bibliography{references}

@article{KO2025113860,
title = {VS-PINN: A fast and efficient training of physics-informed neural networks using variable-scaling methods for solving PDEs with stiff behavior},
journal = {Journal of Computational Physics},
volume = {529},
pages = {113860},
year = {2025},
issn = {0021-9991},
url = {https://www.sciencedirect.com/science/article/pii/S0021999125001433},
author = {Seungchan Ko and Sanghyeon Park}
}

@article{DWIVEDI202096,
title = {Physics Informed Extreme Learning Machine (PIELM)–A rapid method for the numerical solution of partial differential equations},
journal = {Neurocomputing},
volume = {391},
pages = {96-118},
year = {2020},
issn = {0925-2312},
url = {https://www.sciencedirect.com/science/article/pii/S0925231219318144},
author = {Vikas Dwivedi and Balaji Srinivasan}
}

@article{DWIVEDI2025130924,
title = {Curriculum Learning-Driven PIELMs for Fluid Flow Simulations},
journal = {Neurocomputing},
pages = {130924},
year = {2025},
issn = {0925-2312},
url = {https://doi.org/10.1016/j.neucom.2025.130924},
author = {Vikas Dwivedi and Bruno Sixou and Monica Sigovan}
}

@article{CALABRO2021114188,
title = {Extreme learning machine collocation for the numerical solution of elliptic PDEs with sharp gradients},
journal = {Computer Methods in Applied Mechanics and Engineering},
volume = {387},
pages = {114188},
year = {2021},
issn = {0045-7825},
url = {https://www.sciencedirect.com/science/article/pii/S0045782521005193},
author = {Francesco Calabrò and Gianluca Fabiani and Constantinos Siettos}
}

@article{DEFLORIO2024115396,
title = {Physics-Informed Neural Networks for 2nd order ODEs with sharp gradients},
journal = {Journal of Computational and Applied Mathematics},
volume = {436},
pages = {115396},
year = {2024},
issn = {0377-0427},
url = {https://www.sciencedirect.com/science/article/pii/S0377042723003400},
author = {Mario {De Florio} and Enrico Schiassi and Francesco Calabrò and Roberto Furfaro}
}

@article{SCHIASSI2021334,
title = {Extreme theory of functional connections: A fast physics-informed neural network method for solving ordinary and partial differential equations},
journal = {Neurocomputing},
volume = {457},
pages = {334-356},
year = {2021},
issn = {0925-2312},
url = {https://www.sciencedirect.com/science/article/pii/S0925231221009140},
author = {Enrico Schiassi and Roberto Furfaro and Carl Leake and Mario {De Florio} and Hunter Johnston and Daniele Mortari}
}

@Article{Moseley2023,
author={Moseley, Ben
and Markham, Andrew
and Nissen-Meyer, Tarje},
title={Finite basis physics-informed neural networks (FBPINNs): a scalable domain decomposition approach for solving differential equations},
journal={Advances in Computational Mathematics},
year={2023},
month={Jul},
day={31},
volume={49},
number={4},
pages={62},
issn={1572-9044},
doi={10.1007/s10444-023-10065-9},
url={https://doi.org/10.1007/s10444-023-10065-9}
}

@article{SUKUMAR2022114333,
title = {Exact imposition of boundary conditions with distance functions in physics-informed deep neural networks},
journal = {Computer Methods in Applied Mechanics and Engineering},
volume = {389},
pages = {114333},
year = {2022},
issn = {0045-7825},
doi = {https://doi.org/10.1016/j.cma.2021.114333},
url = {https://www.sciencedirect.com/science/article/pii/S0045782521006186},
author = {N. Sukumar and Ankit Srivastava}
}

@misc{dwivedi2025kerneladaptivepielmsforwardinverse,
      title={Kernel-Adaptive PI-ELMs for Forward and Inverse Problems in PDEs with Sharp Gradients}, 
      author={Vikas Dwivedi and Balaji Srinivasan and Monica Sigovan and Bruno Sixou},
      year={2025},
      eprint={2507.10241},
      archivePrefix={arXiv},
      primaryClass={cs.LG},
      url={https://arxiv.org/abs/2507.10241}, 
}

@article{ABBASI2025131440,
title = {Challenges and advancements in modeling shock fronts with physics-informed neural networks: A review and benchmarking study},
journal = {Neurocomputing},
volume = {657},
pages = {131440},
year = {2025},
issn = {0925-2312},
doi = {https://doi.org/10.1016/j.neucom.2025.131440},
url = {https://www.sciencedirect.com/science/article/pii/S0925231225021125},
author = {Jassem Abbasi and Ameya D. Jagtap and Ben Moseley and Aksel Hiorth and Pål Østebø Andersen}
}

@Article{Luo2025,
author={Luo, Kuang
and Zhao, Jingshang
and Wang, Yingping
and Li, Jiayao
and Wen, Junjie
and Liang, Jiong
and Soekmadji, Henry
and Liao, Shaolin},
title={Physics-informed neural networks for PDE problems: a comprehensive review},
journal={Artificial Intelligence Review},
year={2025},
month={Jul},
day={24},
volume={58},
number={10},
pages={323},
issn={1573-7462},
doi={10.1007/s10462-025-11322-7},
url={https://doi.org/10.1007/s10462-025-11322-7}
}

@article{DONG2021114129,
title = {Local extreme learning machines and domain decomposition for solving linear and nonlinear partial differential equations},
journal = {Computer Methods in Applied Mechanics and Engineering},
volume = {387},
pages = {114129},
year = {2021},
issn = {0045-7825},
url = {https://www.sciencedirect.com/science/article/pii/S0045782521004606},
author = {Suchuan Dong and Zongwei Li}
}

@article{RAISSI2019686,
title = {Physics-informed neural networks: A deep learning framework for solving forward and inverse problems involving nonlinear partial differential equations},
journal = {Journal of Computational Physics},
volume = {378},
pages = {686-707},
year = {2019},
issn = {0021-9991},
url = {https://www.sciencedirect.com/science/article/pii/S0021999118307125},
author = {M. Raissi and P. Perdikaris and G.E. Karniadakis}
}

@article{WANG2022110768,
title = {When and why PINNs fail to train: A neural tangent kernel perspective},
journal = {Journal of Computational Physics},
volume = {449},
pages = {110768},
year = {2022},
issn = {0021-9991},
url = {https://www.sciencedirect.com/science/article/pii/S002199912100663X},
author = {Sifan Wang and Xinling Yu and Paris Perdikaris}
}

@article{Gie21092024,
author = {Gung-Min Gie and Youngjoon Hong and Chang-Yeol Jung and},
title = {Semi-analytic PINN methods for singularly perturbed boundary value problems},
journal = {Applicable Analysis},
volume = {103},
number = {14},
pages = {2554--2571},
year = {2024},
publisher = {Taylor \& Francis},
URL = { https://doi.org/10.1080/00036811.2024.2302405}
}

@InProceedings{pmlr-v97-rahaman19a,
  title = 	 {On the Spectral Bias of Neural Networks},
  author =       {Rahaman, Nasim and Baratin, Aristide and Arpit, Devansh and Draxler, Felix and Lin, Min and Hamprecht, Fred and Bengio, Yoshua and Courville, Aaron},
  booktitle = 	 {Proceedings of the 36th International Conference on Machine Learning},
  pages = 	 {5301--5310},
  year = 	 {2019},
  editor = 	 {Chaudhuri, Kamalika and Salakhutdinov, Ruslan},
  volume = 	 {97},
  series = 	 {Proceedings of Machine Learning Research},
  month = 	 {09--15 Jun},
  publisher =    {PMLR},
  url = 	 {https://proceedings.mlr.press/v97/rahaman19a.html}
}

@inproceedings{NEURIPS2020_55053683,
 author = {Tancik, Matthew and Srinivasan, Pratul and Mildenhall, Ben and Fridovich-Keil, Sara and Raghavan, Nithin and Singhal, Utkarsh and Ramamoorthi, Ravi and Barron, Jonathan and Ng, Ren},
 booktitle = {Advances in Neural Information Processing Systems},
 editor = {H. Larochelle and M. Ranzato and R. Hadsell and M.F. Balcan and H. Lin},
 pages = {7537--7547},
 publisher = {Curran Associates, Inc.},
 title = {Fourier Features Let Networks Learn High Frequency Functions in Low Dimensional Domains},
 url = {https://proceedings.neurips.cc/paper_files/paper/2020/file/55053683268957697aa39fba6f231c68-Paper.pdf},
 volume = {33},
 year = {2020}
}

@Article{CiCP-28-5,
author = {Zhi-Qin, Xu, John and Zhang, Yaoyu and Tao, Luo and Yanyang, Xiao and Zheng, Ma},
title = {Frequency Principle: Fourier Analysis Sheds Light on Deep Neural Networks},
journal = {Communications in Computational Physics},
year = {2020},
volume = {28},
number = {5},
pages = {1746--1767},
issn = {1991-7120},
url = {https://global-sci.com/article/79739/frequency-principle-fourier-analysis-sheds-light-on-deep-neural-networks}
}

@article{ARZANI2023111768,
title = {Theory-guided physics-informed neural networks for boundary layer problems with singular perturbation},
journal = {Journal of Computational Physics},
volume = {473},
pages = {111768},
year = {2023},
issn = {0021-9991},
url = {https://www.sciencedirect.com/science/article/pii/S0021999122008312},
author = {Amirhossein Arzani and Kevin W. Cassel and Roshan M. D'Souza}
}

@Article{Klawonn2024,
author={Klawonn, Axel
and Lanser, Martin
and Weber, Janine},
title={Machine learning and domain decomposition methods - a survey},
journal={Computational Science and Engineering},
year={2024},
month={Sep},
day={23},
volume={1},
number={1},
pages={2},
issn={2948-1597},
doi={10.1007/s44207-024-00003-y},
url={https://doi.org/10.1007/s44207-024-00003-y}
}

@article{DWIVEDI_DPINN_2021,
title = {Distributed learning machines for solving forward and inverse problems in partial differential equations},
journal = {Neurocomputing},
volume = {420},
pages = {299-316},
year = {2021},
issn = {0925-2312},
url = {https://doi.org/10.1016/j.neucom.2020.09.006, https://arxiv.org/abs/1907.08967},
author = {Vikas Dwivedi and Nishant Parashar and Balaji Srinivasan}
}

@article{10.1093/imamat/hxae011,
    author = {Grossmann, Tamara G and Komorowska, Urszula Julia and Latz, Jonas and Schönlieb, Carola-Bibiane},
    title = {Can physics-informed neural networks beat the finite element method?},
    journal = {IMA Journal of Applied Mathematics},
    volume = {89},
    number = {1},
    pages = {143-174},
    year = {2024},
    month = {05},
    issn = {0272-4960},
    doi = {10.1093/imamat/hxae011},
    url = {https://doi.org/10.1093/imamat/hxae011},
    eprint = {https://academic.oup.com/imamat/article-pdf/89/1/143/58325885/hxae011.pdf},
}

@misc{ren2025generalfourierfeaturephysicsinformed,
      title={General Fourier Feature Physics-Informed Extreme Learning Machine (GFF-PIELM) for High-Frequency PDEs}, 
      author={Fei Ren and Sifan Wang and Pei-Zhi Zhuang and Hai-Sui Yu and He Yang},
      year={2025},
      eprint={2510.12293},
      archivePrefix={arXiv},
      primaryClass={cs.LG},
      url={https://arxiv.org/abs/2510.12293}, 
}

\end{document}